\documentclass{bmvc2k}
\title{Detecting Reflections by Combining Semantic and Instance Segmentation}

\addauthor{David Owen}{david.owen@umbocv.com}{1}
\addauthor{Ping-Lin Chang}{ping-lin.chang@umbocv.com}{1}

\addinstitution{
Umbo Computer Vision
}

\runninghead{Owen, Chang}{Detecting Reflections}
\begin{document}

\maketitle
%\thispagestyle{empty}

%%%%%%%%% ABSTRACT
\begin{abstract}
	Reflections in natural images commonly cause false positives in automated detection systems. These false positives can lead to significant impairment of accuracy in the tasks of detection, counting and segmentation. Here, inspired by the recent \textit{panoptic} approach to segmentation, we show how fusing instance and semantic segmentation can automatically identify reflection false positives, without explicitly needing to have the reflective regions labelled. We explore in detail how state of the art two-stage detectors suffer a loss of broader contextual features, and hence are unable to learn to ignore these reflections. We then present an approach to fuse instance and semantic segmentations for this application, and subsequently show how this reduces false positive detections in a real world surveillance data with a large number of reflective surfaces. This demonstrates how panoptic segmentation and related work, despite being in its infancy, can already be useful in real world computer vision problems.
\end{abstract}

%%%%%%%%% BODY TEXT
\section{Introduction}
Reflective surfaces are common in natural images, and can cause false positive results in the tasks of detection, segmentation and counting~\cite{bajcsy1996detection,delpozo2007detecting,braun2018eurocity}. In a setting with many reflective surfaces, such as a gym with mirrors on the walls, the scale of these false positives can be significant. For instance, in a room with one large mirror along the wall and several people next to it, there may be as many false positive detections (due to reflections) as there are true detections (due to people).

\begin{figure}[h]
	\centering
	\includegraphics[width=0.32\textwidth]{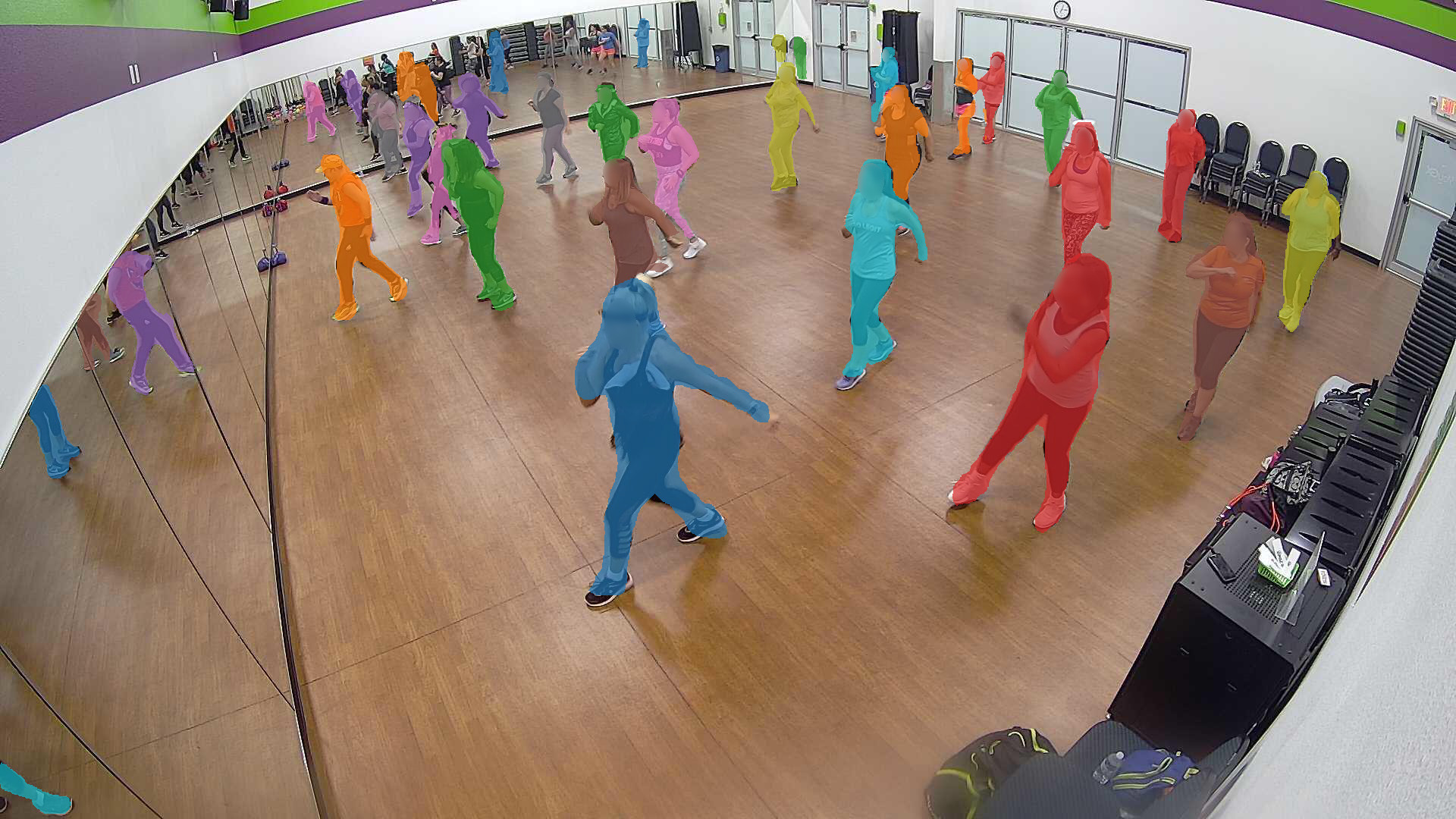}
	\includegraphics[width=0.32\textwidth]{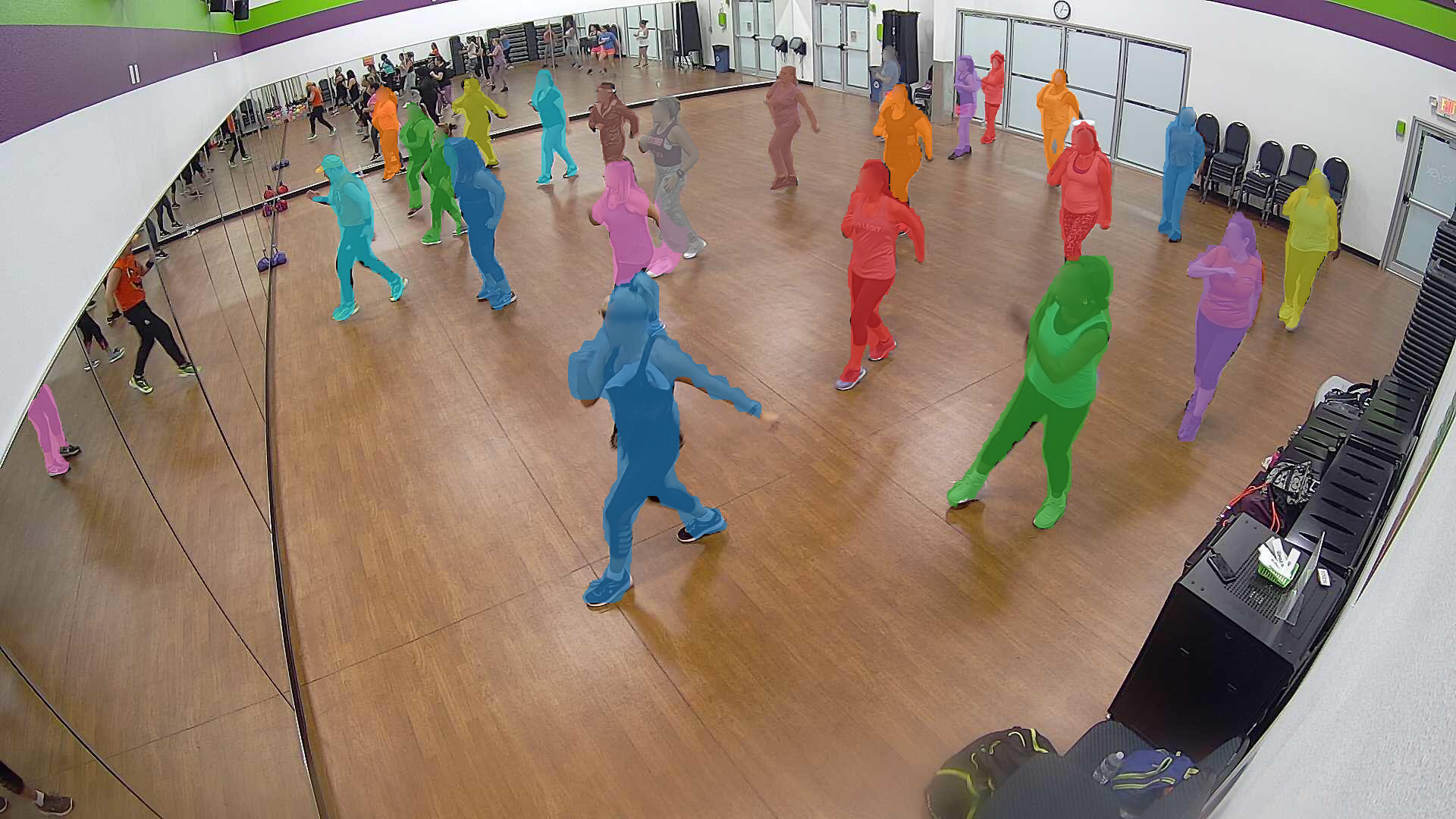}
	\includegraphics[width=0.32\textwidth]{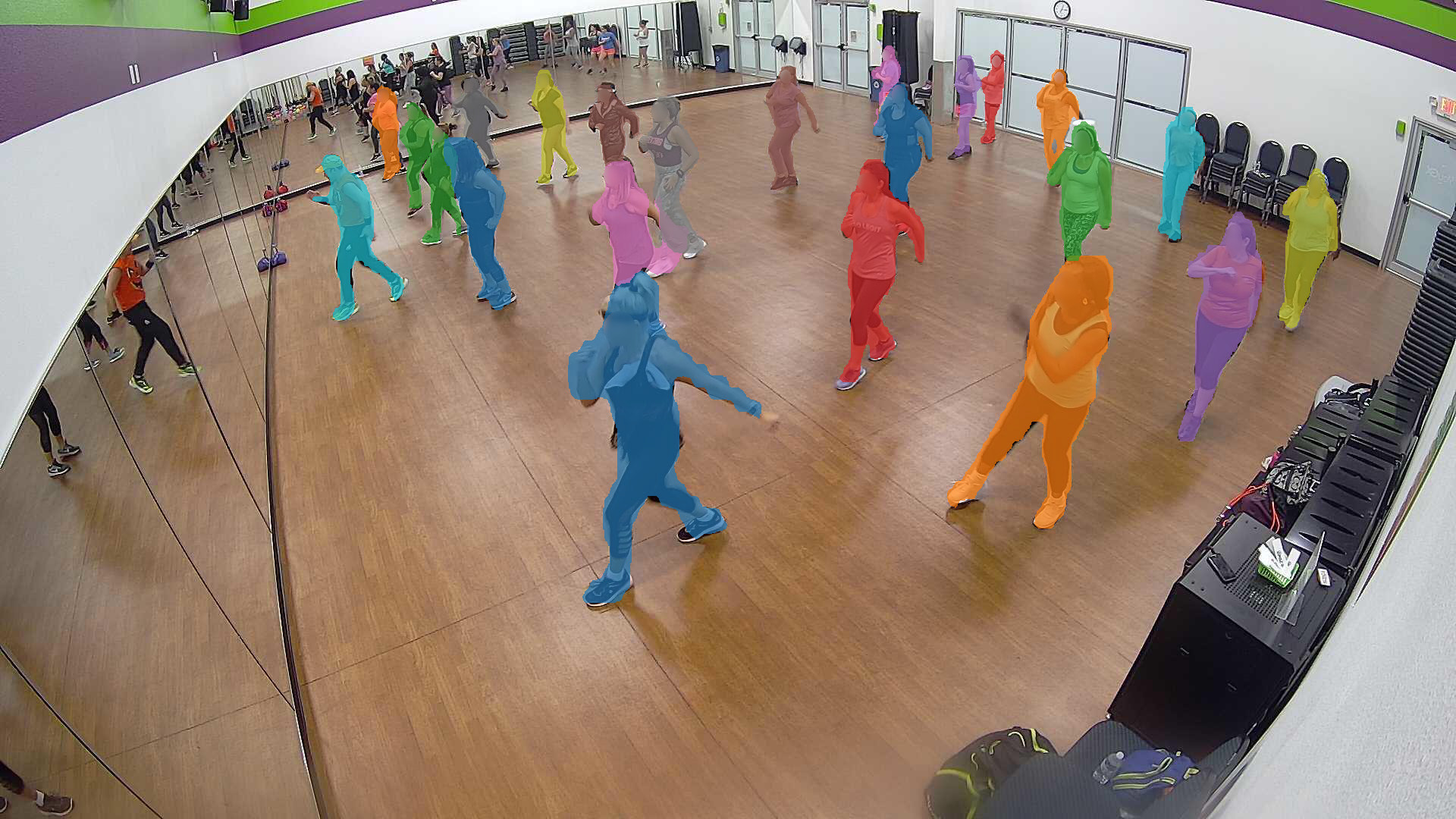}
	\caption{Example images with instance segmentation. \textbf{Left}: Mask RCNN trained on a dataset without reflections; \textbf{Middle}: Mask RCNN trained on a dataset with reflections; \textbf{Right}: Joint segmentation as in this work. Instance segmentation struggles to distinguish reflections from true positives, even after finetuning on reflections.}
	\label{fig:examples}
\end{figure}

One might expect that this problem could be addressed by finetuning on a dataset with mirrors -- usually, finetuning allows a pretrained network to adapt to a different domain. In this work, we show this approach is ineffective for two-stage detection/segmentation methods such as the popular Mask RCNN (see Figure~\ref{fig:examples}), and that this is due to the loss of broader image context during the second stage. We proceed to show that semantic segmentation methods are unaffected by this loss of context, and are better able to learn to distinguish reflections from true positives. Subsequently, inspired by the recently introduced concept of \textit{panoptic segmentation} (i.e. a fusion of instance and semantic segmentation), we show how false positive detections due to reflections can be dramatically reduced while retaining instance-level segmentations, invaluable for tasks such as counting and tracking.

While this work is limited to mirror reflections, it shows the promise of panoptic segmentation and related approaches, revealing weaknesses in the popular two-stage detection approaches and how these can easily be addressed through the introduction of broader context information as is retained better in typical semantic segmentation architectures. This points the way towards other promising areas where panoptic segmentation may improve performance, such as segmentation of small objects and other tasks where the high-level information of \textit{scene parsing} is important for understanding fine-grained components of the scene. Indeed, this work shows how even at this very early stage, combining instance and semantic segmentation can be practically valuable in real world computer vision problems.

%One way to address this problem is to manually label images with the location of reflective surfaces. However, in addition to requiring effort on the part of the labeller, 

\section{Related Work}
\paragraph*{Instance Segmentation} Given an input image, instance segmentation returns a set of detected objects and their segmentations. Instance segmentation focuses on the detection and segmentation of \textit{things} -- specific and distinct instances of objects, such as people and cars. Instance segmentation is most popularly done using two-stage methods such as Mask RCNN, which first use a region proposal network (RPN) to identify regions that potentially contain instances, and subsequently uses a classification stage to decide: (i) whether the region contains an instance; (ii) where the instance is within the region; (iii) which specific pixels belong to the instance. The two-stage architecture gives state of the art results for detection and segmentation, at the cost of greater computational (and conceptual) complexity~\cite{huang2017speed}. Single-stage architectures such as RetinaNet~\cite{lin2017focal} have attempted to achieve similar results, particularly in detection, but currently are weaker.

\paragraph*{Semantic Segmentation} In contrast to instance segmentation, semantic segmentation aims to produce a label for each pixel in the image. Semantic segmentation focuses on the segmentation of \textit{stuff} -- there is no importance attached to distinguishing different instances of a class, but rather to generally recognising the composition of a scene. Typically, semantic segmentation uses a simpler single-stage architecture, with the state of the art architectures designed to preserve information across a range of feature scales.

Crucially, because semantic segmentation is done in a single stage, broader-scale contextual information is usually available for the detection and segmentation tasks~\cite{fu2018dual,chen2019hybrid}. Research on the \textit{effective receptive field} has shown that, when estimating the class label for a given point in an image, the network gives highest priority to the features derived from nearby pixels but is able to be significantly influenced by pixels from further away~\cite{peng2017large,luo2016understanding}. Various techniques may encourage the segmentation approach to incorporate large-scale dependencies across the image by directly expanding the effective receptive field through network structure or purpose-created modules~\cite{chen2019hybrid,fu2018dual,chen2018deeplab}, attempting to fuse multiple scales of visual information~\cite{zhao2017pyramid,zhang2018context}, or using auxiliary models of context, e.g. using LSTM~\cite{byeon2015scene,shuai2018scene}. This contrasts with two-stage instance segmentation where, as we show in Section~\ref{sec:methods}, contextual information is necessarily lost after the RPN stage.

\paragraph*{Panoptic Segmentation} The recently proposed idea of panoptic segmentation has gathered much interest, both as a unifying set of metrics in which to compare semantic and instance segmentation, and as a motivating problem that may lead to novel segmentation architectures~\cite{kirillov2018panoptic}. In panoptic segmentation, every pixel in the image should be labelled, but different instances are distinguished from one another, e.g. ``person 1'' is distinguished from ``person 2''. Human-level performance on panoptic segmentation, or full scene parsing, has been described as a ``Holy Grail'' of computer vision~\cite{zhou2017scene}. Intuitively, it is appealing to unify the two fundamentally related tasks of semantic and instance segmentation, which have historically been treated separately~\cite{kirillov2018panoptic}.

Thus far, most approaches to panoptic segmentation have used two separate segmentation methods, with each being separately trained, and their proposed segmentations being fused only at the end of each pipeline. This fusion can be performed algorithmically, as in the original presentation of panoptic segmentation and the panoptic segmentation challenge~\cite{kirillov2018panoptic}. Alternatively, the fusion may be performed as its own subnetwork within the network architecture~\cite{de2018panoptic}. Most recently, unified architectures are beginning to emerge, although currently these perform significantly worse when trained end-to-end than using the aforementioned fusion approaches on separately trained networks~\cite{kirillov2019panoptic,li2018attention,de2018panoptic,xiong19upsnet}.

We do not call the approach developed in this work \textit{panoptic}, as rather than focusing on a detailed multi-class annotation of scenes we very specifically focus on detection of two classes: person and background (with the background class including mirror reflections). However, our approach is heavily inspired by the existing work on panoptic segmentation, and arguably any segmentation approach that fuses semantic and instance information may be seen as a form of panoptic segmentation.

\paragraph*{Detecting Reflections} There are few existing solutions to automatically detect reflections in an image. Typically, in an applied computer vision context, problematic mirrors might be ignored through the use of a hand-drawn ROI. This, of course, requires human effort to label the ROI, and depends upon the camera and mirror remaining in a fixed perspective. There has been some work on the automatic detection of reflecting planes using geometric models of the image and its reflection~\cite{delpozo2007detecting,bajcsy1996detection}, but this was not explored in the context of segmentation, and had several barriers to practical use.

Much of the recent work on mirror detection has been motivated by autonomous vehicle research, and has been in the context of simultaneous localisation and mapping (SLAM) using different sensors~\cite{aggarwal2016detection}, or active projection~\cite{kashammer2015mirror}, rather than focusing on colour images. Work on colour images in vehicle-heavy scenes has argued that, particularly for classification error, ``depictions and reflections are among the hardest error sources to take care of''~\cite{braun2018eurocity}. Separately, such work makes the case for manual annotation of difficult image regions as ``ignore'' regions that should be excluded from the RPN during training, although this is not explored for reflections specifically, and would again require manual annotation~\cite{braun2018eurocity,cordts2016cityscapes}.

\section{Learning to Ignore Reflections}
\label{sec:methods}
\begin{figure}[h]
	\centering
	\includegraphics[height=7.95em]{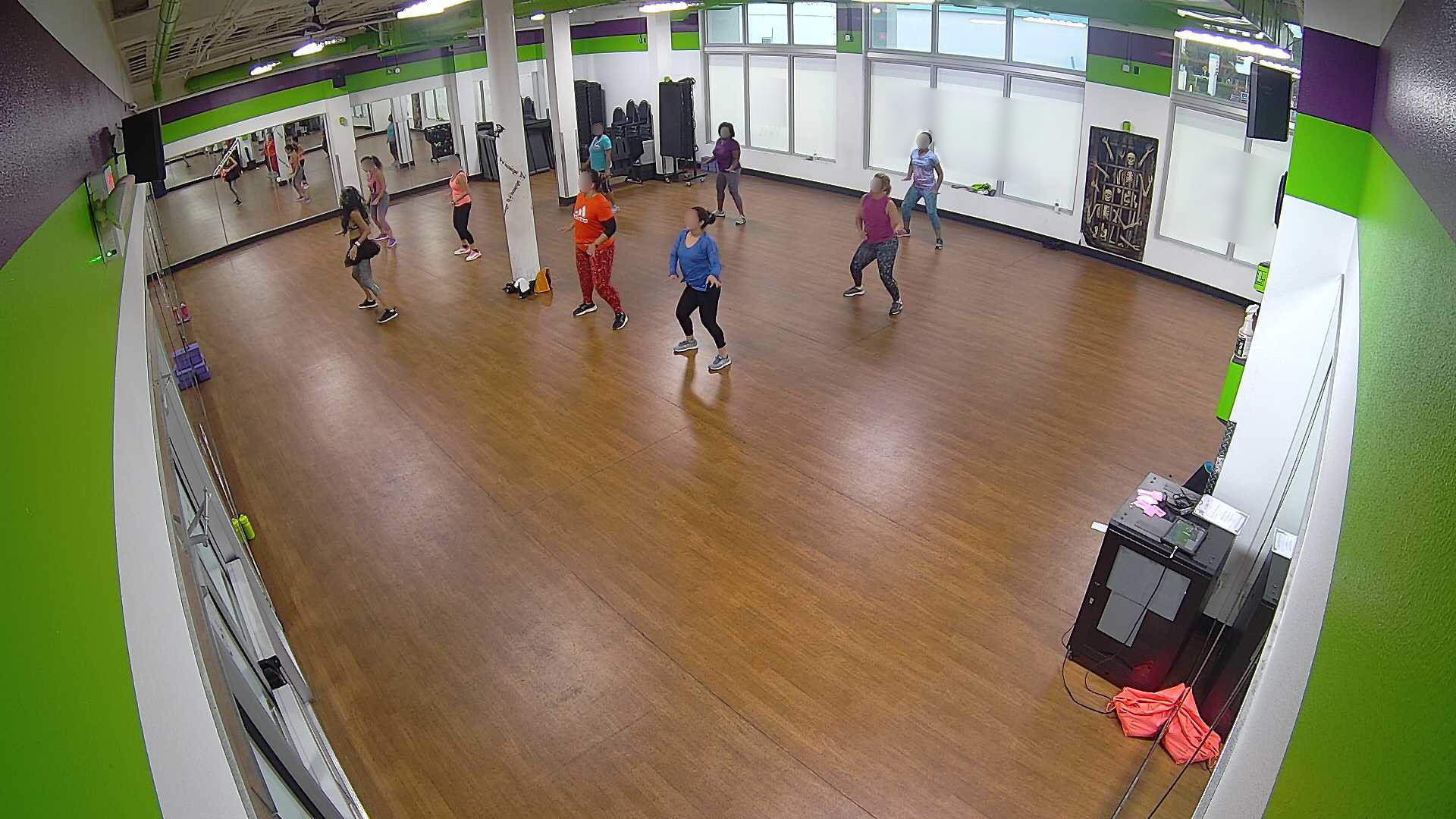}
	\includegraphics[height=7.95em]{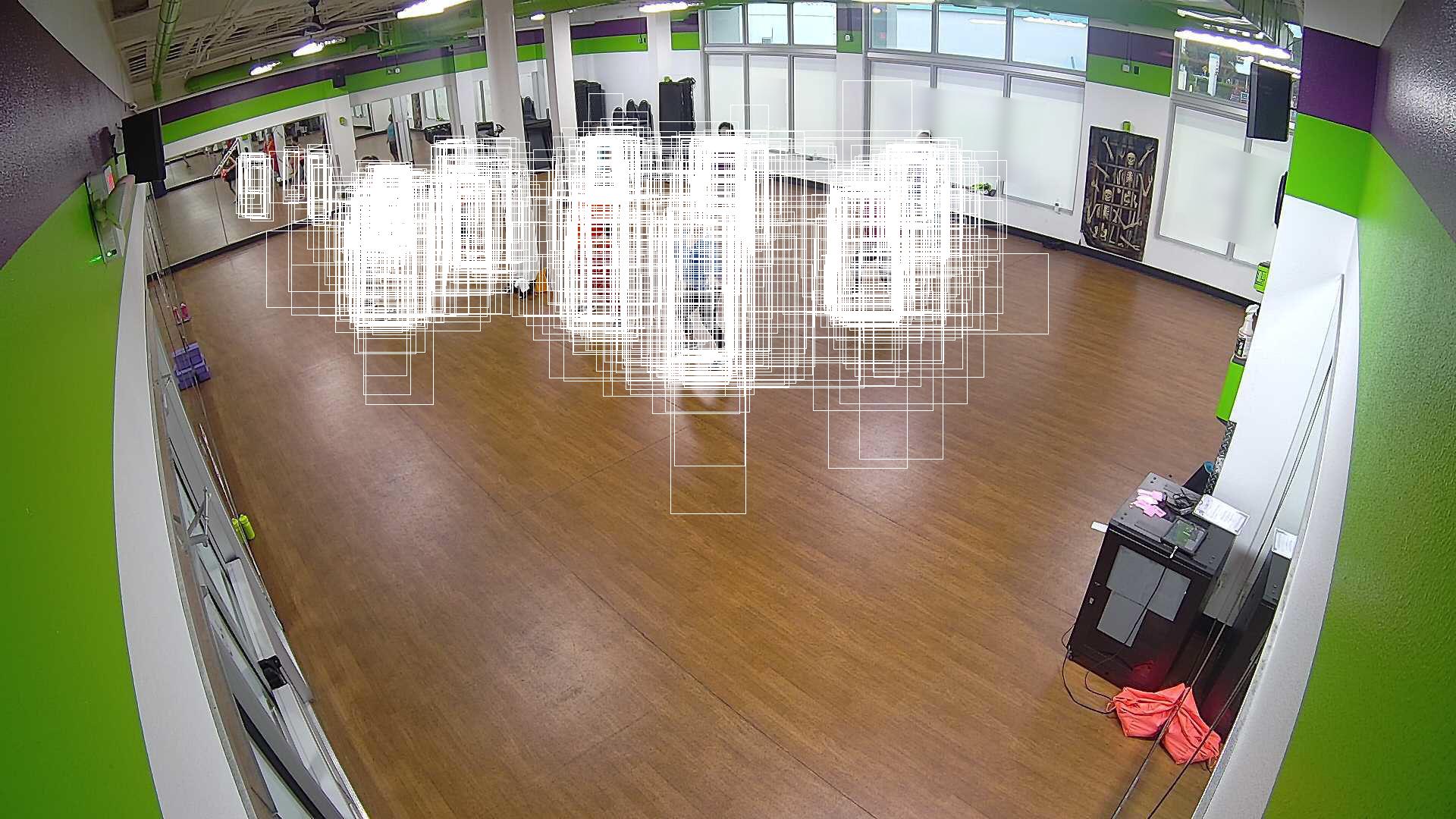}
	\includegraphics[height=7.95em]{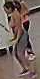}
	\includegraphics[height=7.95em]{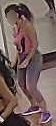}
	\caption{From left: example image, RPN proposals, example reflection, example true positive. In the classification stage, broader context around the proposal is discarded, and detecting whether a proposal is a reflection becomes challenging or even impossible. However it is difficult to train the RPN stage to completely ignore all mirror regions.}
	\label{fig:rpn}
\end{figure}
A general solution to the problem of reflections would require an advanced understanding of scene composition and light properties. Rather than attempting to model this from first principles, as in previous work on the topic~\cite{delpozo2007detecting}, we note that important practical progress can be made on an easier subproblem: Can we automatically learn to ignore instances in regions that look like mirrors? Modern machine learning approaches might be able to learn to recognise mirror regions based on differences in their visual properties: distortion, brightness, having a distinctive frame around the mirror, and so on.

This motivated us to finetune a state of the art instance segmentation network (Mask RCNN) on a dataset containing several mirrors (details of the dataset are described in Section~\ref{sec:data}). We expected that the network would learn to ignore many of the reflections. This was largely ineffective: Finetuning retained significant numbers of false positive detections. To our surprise, however, a semantic segmentation architecture trained on the same dataset containing several mirrors was able to almost entirely dismiss these false positive detections.

By exploring how the images were assessed step-by-step as they were processed by Mask RCNN, we found the reason for this disparity. There is a non-obvious weakness present in Mask RCNN and other two-stage detectors: Once a region has been proposed by the RPN, broader context outside the proposal is necessarily discarded (see Figure~\ref{fig:rpn}). As a result, the post-RPN classification stage cannot use broader contextual features, such as the frame of the mirror, to recognise that a proposal was a reflection. However, because the RPN is required by design to give a large number of candidate regions, the RPN cannot effectively be trained to avoid proposing regions inside mirrors. That is: Only the RPN can see whether a proposal is inside a mirror, but the RPN is forced to yield a certain number of these proposals nonetheless. Semantic segmentation architectures have no such separation of concerns, and hence are easily able to dismiss reflections.

This inspired us to guide the instance segmentation away from these false positives by using the semantic segmentation results. Segmentation approaches that fuse semantic and instance segmentation -- recently called \textit{panoptic} segmentation -- have demonstrated a promising capability to reduce challenging false positives~\cite{arnab2017pixelwise}. A broader understanding of the scene, as provided by semantic segmentation, can guide instance segmentation and vice versa. We follow previous work on panoptic segmentation by using a straightforward heuristic method to fuse semantic and instance segmentations, producing a joint segmentation~\cite{kirillov2018panoptic}.

In our fusion method we simply examine, for each proposed instance, the average semantic segmentation score within the proposed mask. If $\sum_{I} score$ is greater than some threshold, $c$, then we accept the instance segmentation; otherwise, we reject it. A key difference of our approach lies in how the fusion of semantic and instance segmentations occurs. In the original heuristic approach of He \textit{et al}, conflicts are always resolved in favour of instances. This approach was not only used in their original work, but also in their most recent work on performing panoptic segmentation with a single network. In our approach, conversely, semantic results are allowed to take priority when their score is sufficiently strong.

Intuitively one might expect $c=0.5$ to be optimal, on the rationale that this accepts an instance when semantic segmentation thinks it is, overall, likelier than not. However, this is not the case: Semantic segmentation is worse for detection than state of the art instance segmentation, and setting the threshold high leads to a large increase in false negatives. Moreover, discrepancies between the proposed masks can further weaken the average score. Instead of setting $c=0.5$, we make it a tunable parameter, and tune it on the validation set after training instance and semantic networks. We present these tuning results in Section~\ref{sec:tuning}.

\section{Experiments}
\subsection{Setup}
\label{sec:data}

Experiments in this work are based on a real world dataset containing slightly over 22,000 $1920\times1080$ surveillance images from gyms, with 527 of these manually determined to include a significant degree of reflections. Throughout this work, 21,500 reflection-free images are used for pretraining; and the remaining 527 images are used for finetuning and assessment of mirror false positives. This dataset was collected from an industrial partner, and is used to demonstrate how our method holds up in real data. The surveillance dataset covers several settings, and scenes have a lot of variation in the number of people present, the activities they are doing, and whether or not there are significant reflections. In the most extreme perspectives, the gyms have large mirrors mounted on the walls, and reflections can potentially cause false positives on the order of 100\%, as in Figure~\ref{fig:examples}.

The baseline two-stage detection and instance segmentation method we examine in this work is Mask RCNN without bells and whistles~\cite{massa2018mrcnn,he2017mask}, trained on the COCO dataset~\cite{lin2014microsoft}. For the semantic branch of our joint segmentation we use UPerNet~\cite{zhou2017scene,zhou2018semantic,xiao2018unified} with a ResNet50 backbone, which is inspired by the FPN of Mask RCNN and provides near state of the art results~\cite{xiao2018unified}. UPerNet was pretrained on the ADE20K dataset~\cite{zhou2018semantic,zhou2017scene}.

UPerNet is trained on all images, enabling it to learn to detect mirrors without their being explicitly labelled. In Section~\ref{sec:context}, we examine in detail how UPerNet is able to do this, using edited images to separate the contributions of contextual information, geometric features and textural features. Subsequently, in Section~\ref{sec:tuning} we show how to determine the optimal tuning parameter, $c$, to determine the best balance between instance and semantic results.

Finally, we explore performance in two different cases: In the first case, Mask RCNN is pretrained on images \textit{without} mirrors, and in the second case, the pretrained network is finetuned on images including mirrors. The former approach sets a baseline for assessment, and shows the most drastic degree of false positive detections. The latter approach represents the standard way to attempt to suppress these false positives (finetuning on example images). We show results for Mask RCNN without further processing in each case (non-finetuned and finetuned), and then we show the fusion approach set out in Section~\ref{sec:methods} gives significant improvements in rejecting false positives.
\begin{figure}[h]
	\centering
	\includegraphics[height=10em]{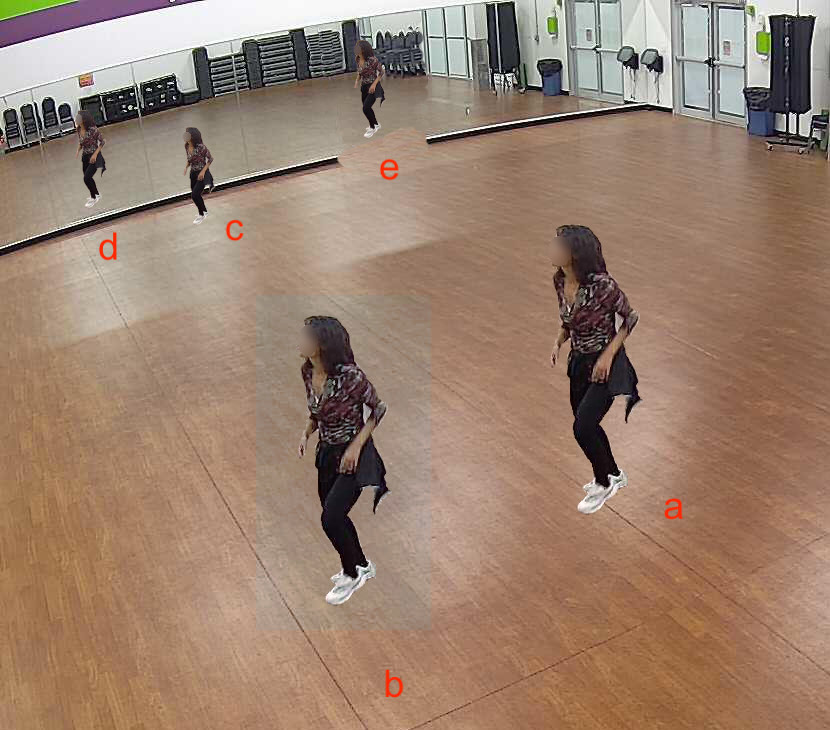}
	\includegraphics[height=10em]{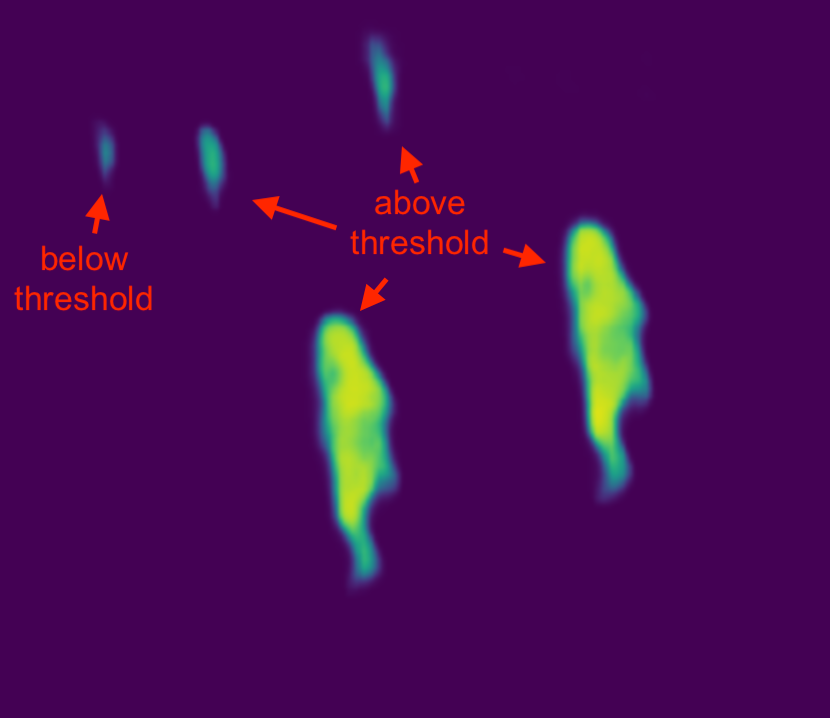}
	\includegraphics[height=10em]{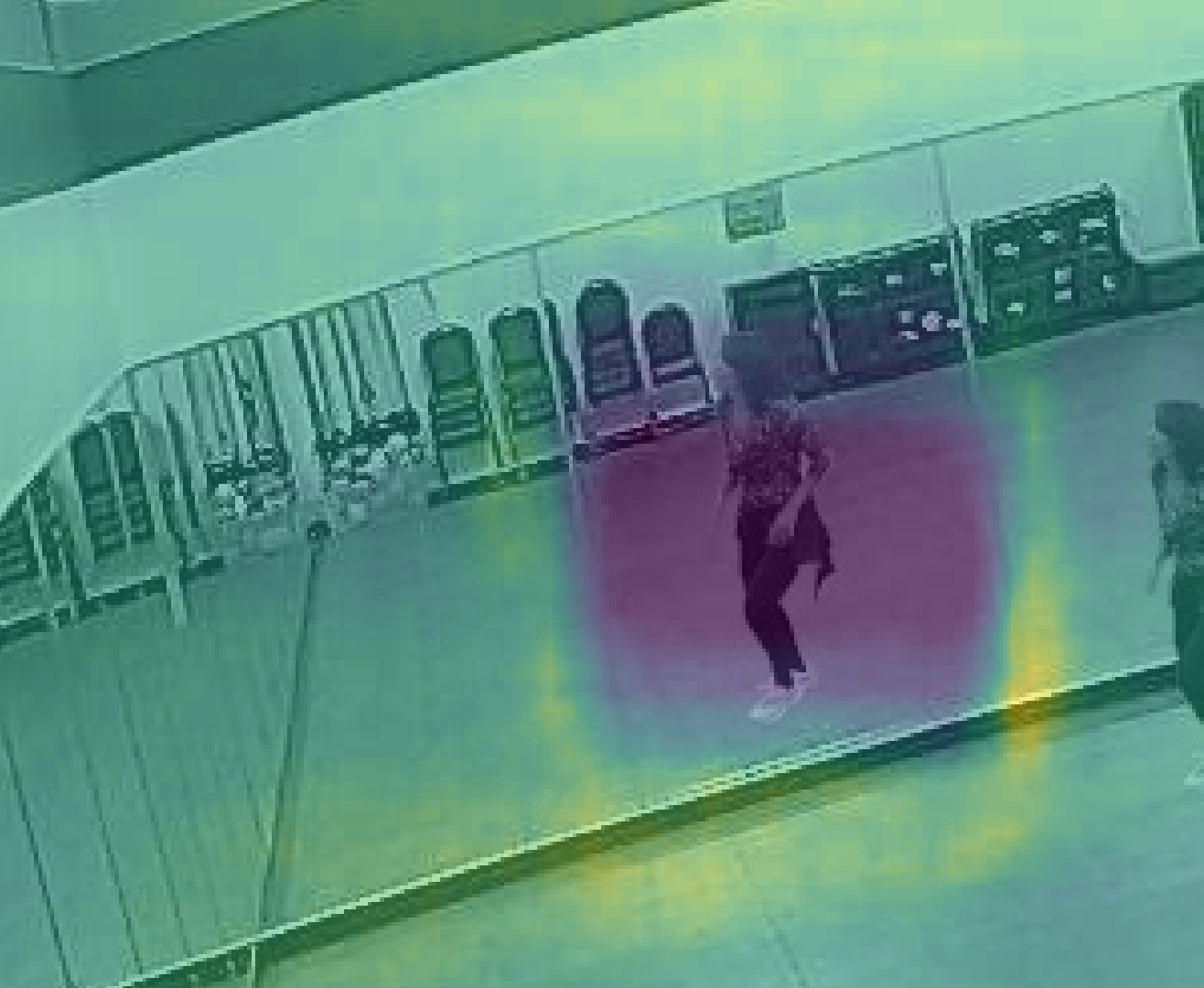}
	\caption{\textbf{Left}: a synthesised image, created to show how contextual cues such as the mirror frame allow semantic segmentation to recognise reflections. \textbf{Middle}: the corresponding segmentation score map. \textbf{Right}: heatmap of how masking image features affects the semantic score of d), evaluated by sliding a $96\times54$ window step size 5 pixels in each direction.}
	\label{fig:edited-demo}
\end{figure}

\subsection{Geometry, Texture or Context?}
\label{sec:context}

The detection of reflection false positives could, in theory, stem from any of the information in the image. This might be from geometry (e.g. reflections are smaller than true positives; also, by definition, reflections mirror the original image), texture (the reflected image might have different textural properties), or scene context (e.g. the frame of the mirror surrounding the reflection reveals that it is a reflection, and not a true positive). Here, we show that the key information for the semantic segmentation lies in contextual cues -- specifically, the mirror frame. To do this, we use photo editing to clone a true positive detection and adjust these properties separately, as shown in Figure~\ref{fig:edited-demo}.

The image contains: a) a regular non-mirror true positive; b) a copy of a), with the color profile changed to reflect that of a mirror reflection; c) a copy of a) downscaled to be the same size as a reflection, placed outside the mirror; d) an identically rescaled copy to c), placed within the mirror; e) an identically rescaled copy to c), placed within the mirror, but with the mirror frame manually erased to blend the mirror region with its surroundings.This allows us to disentangle the effects of texture (a versus b), geometry (c versus d) and context (d versus e). As revealed by the semantic segmentation score map, the only factor that leads to no detection (a true negative) is the context. Only d) is dismissed as a reflection, and e) is not recognised as a reflection due to the removal of the mirror frame. Geometric and textural information remains unchanged from the original true positive, meaning the only change is the contextual information from the frame.

We show the effects of the mirror frame in greater detail using an established visualisation technique~\cite{zeiler2014visualizing} in which a grey patch is used to block out parts of the image before segmentation. We then evaluate the segmentation score, averaged over the visible ground truth mask. This gives a map showing how different regions affect detection. Matching intuitive expectations, blocking out regions containing the person dramatically reduces the segmentation score, while blocking out parts of the floor has a negligible effect. More importantly, however, blocking out parts of the \emph{frame} increases the segmentation score -- confirming that the frame is key to detecting reflections.

\subsection{Balancing Instance and Semantic Segmentation}
\label{sec:tuning}
\begin{figure}
	\centering
	\begin{subfigure}{1.0\textwidth}
	\includegraphics[width=0.32\textwidth]{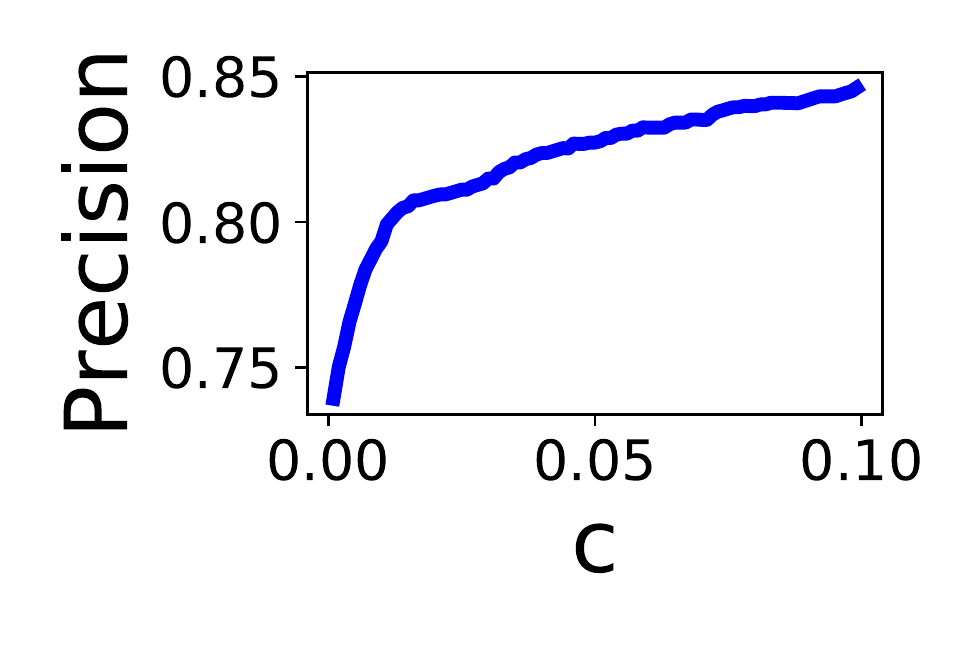}
	\includegraphics[width=0.32\textwidth]{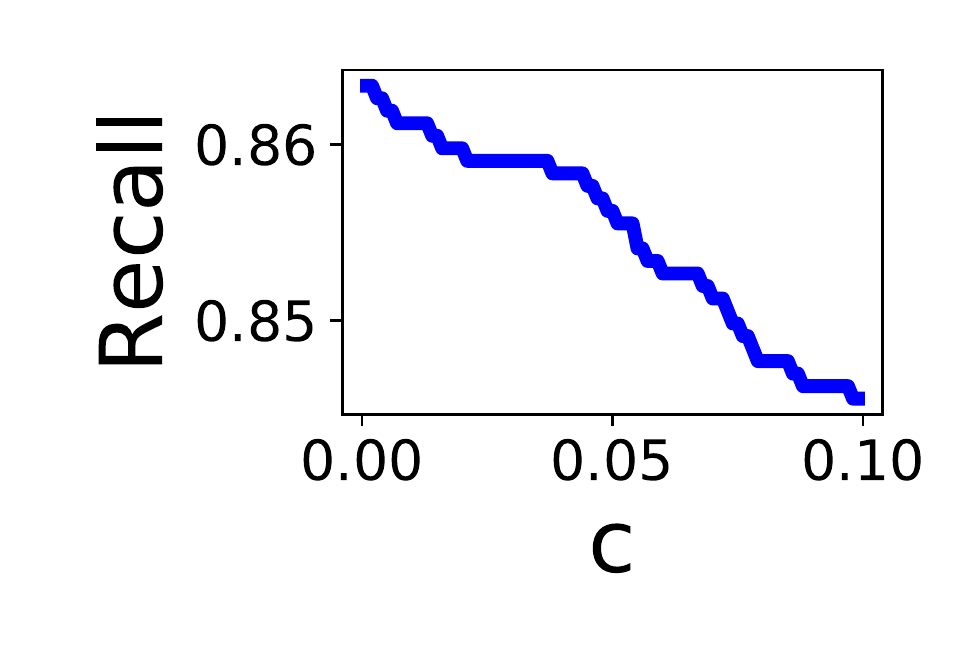}
	\includegraphics[width=0.32\textwidth]{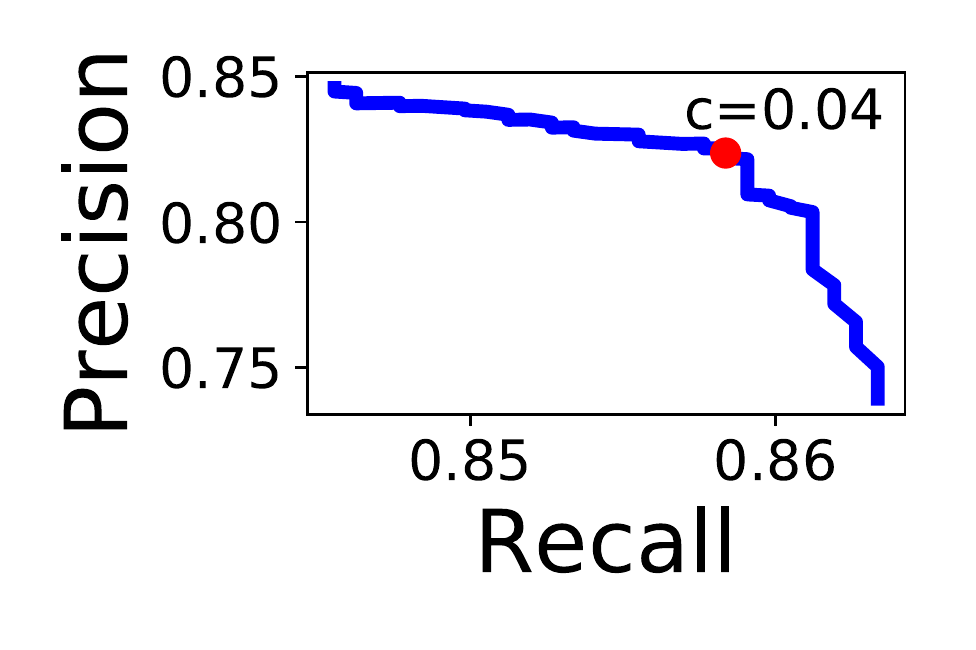}
		\caption{Surveillance data without reflections.}
	\end{subfigure}
	\begin{subfigure}{1.0\textwidth}
	\includegraphics[width=0.32\textwidth]{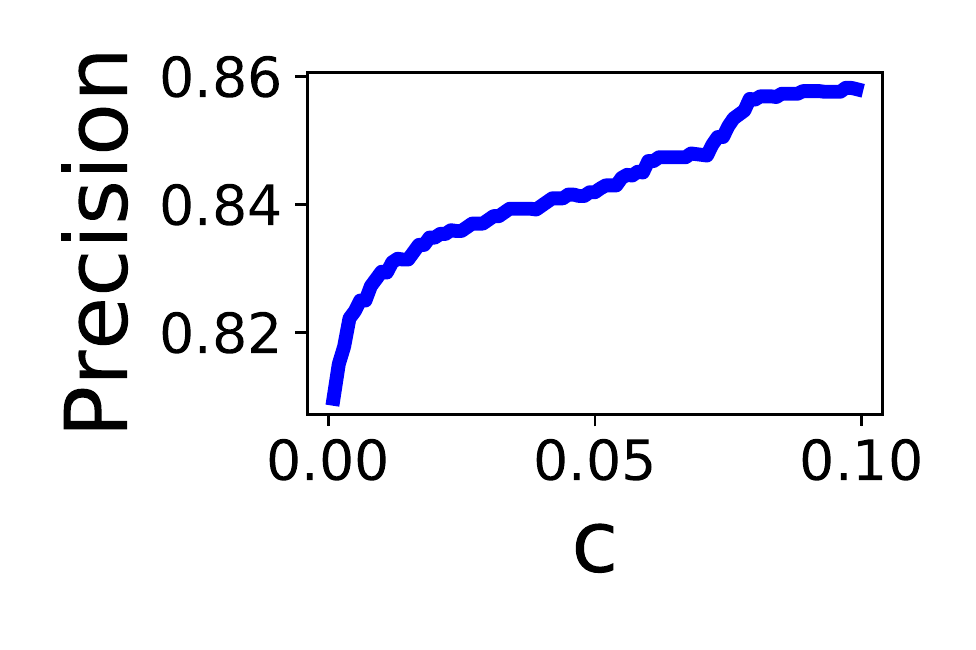}
	\includegraphics[width=0.32\textwidth]{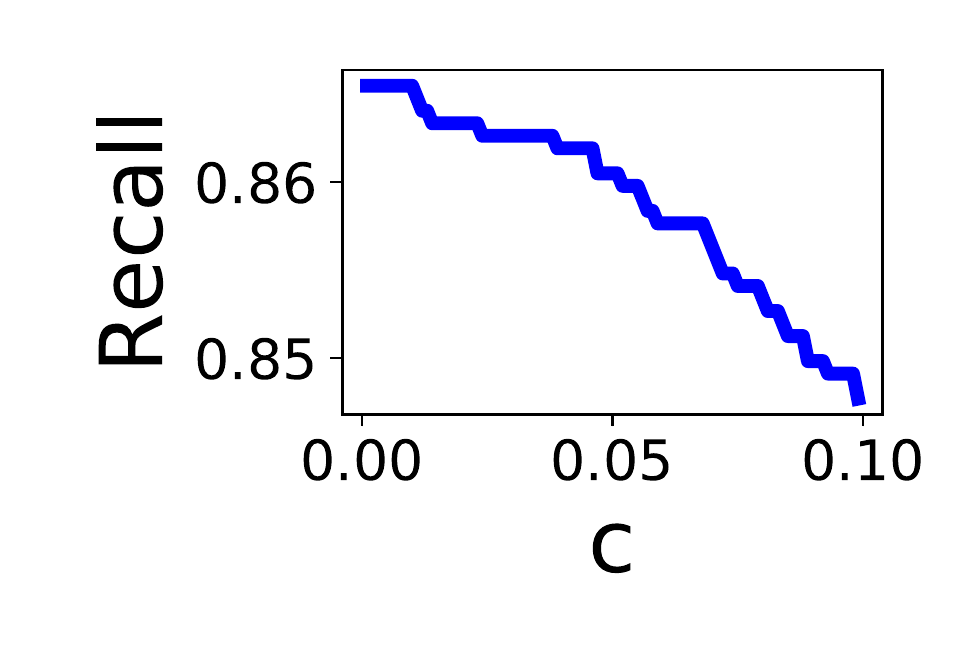}
	\includegraphics[width=0.32\textwidth]{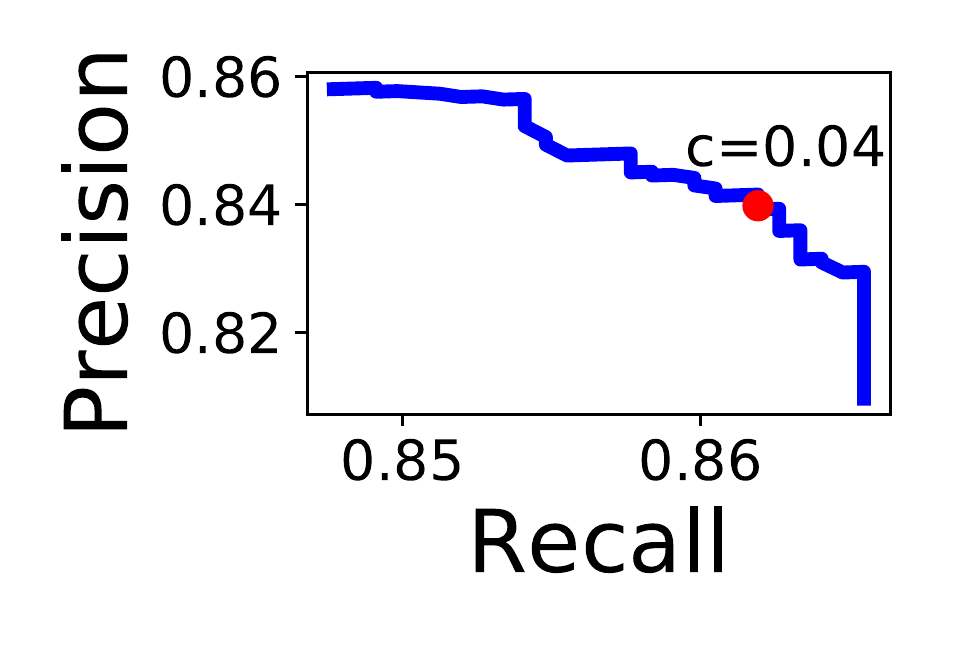}
		\caption{Finetuned on mirror reflections.}
	\end{subfigure}
	\vspace{1em}
	\caption{Tuning the semantic threshold parameter, $c$. For each network, from left to right: precision versus $c$,  recall versus $c$, precision versus recall for different values of $c$.}
	\label{fig:tuning}
\end{figure}

As discussed in Section~\ref{sec:methods}, our fusion approach introduces a parameter, $c$, controlling the trade-off between semantic and instance segmentation. As $c$ is increased, there are stricter requirements from semantic predictions: More false positives will be removed, at the cost of increasing false negative predictions. In order to find the optimal trade-off, we examine how $c$ affects precision and recall in the validation set, both for the pretrained and the pretrained+finetuned Mask RCNN networks. These results are shown in Figure~\ref{fig:tuning}.

We settle on $c=0.04$, which yields appreciable improvements in precision without greatly compromising recall. This is clearest in the precision-recall curves, which show how $c=0.04$ is roughly at the top-right ``knee'' of the curve. This is particularly visible in the pretrained case. Based on these results, for all subsequent experiments we use $c=0.04$.

\subsection{Performance}
\begin{figure}[h]
	\centering
	\begin{subfigure}{\textwidth}
	\includegraphics[width=0.19\textwidth]{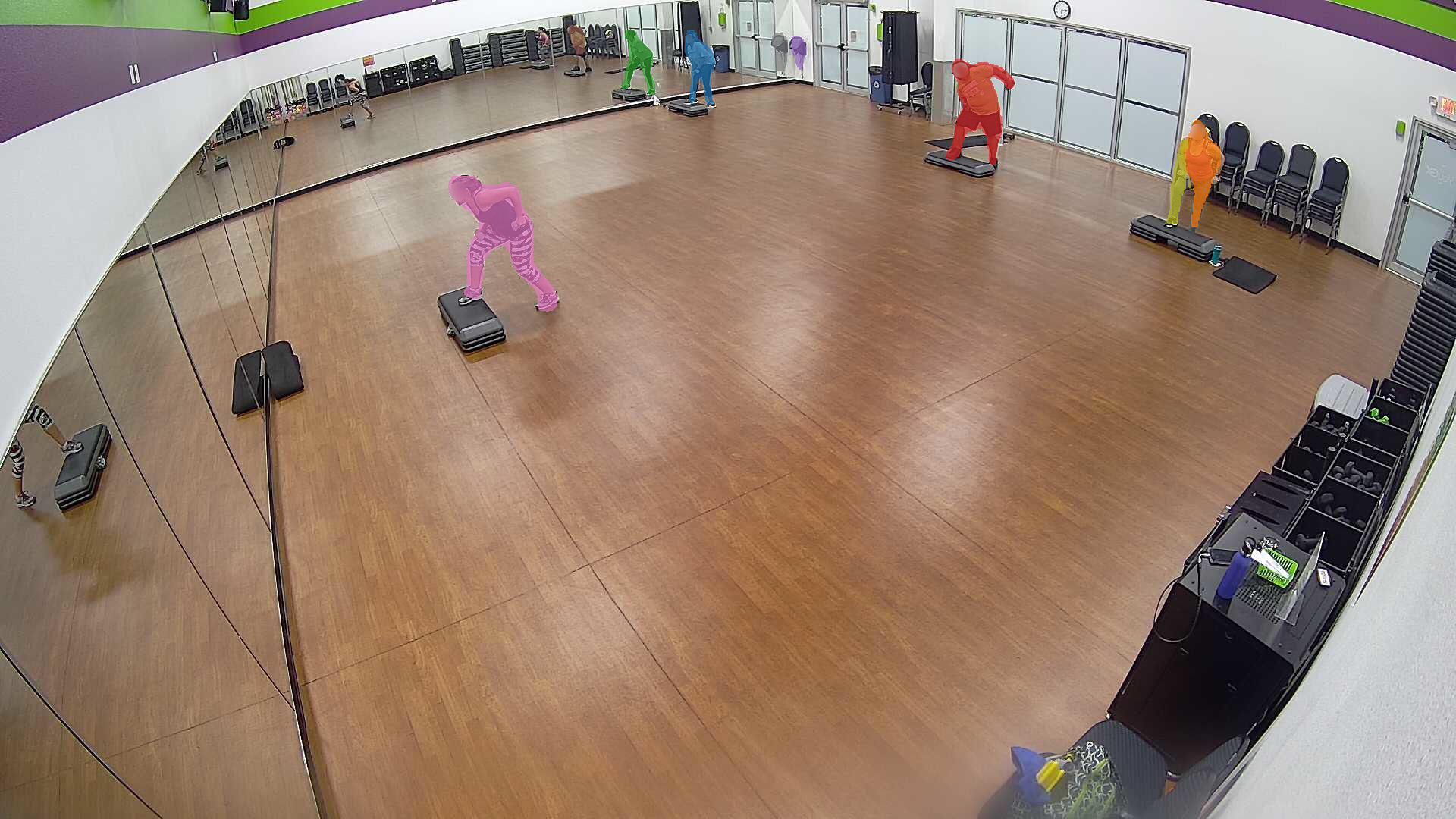}
	\includegraphics[width=0.19\textwidth]{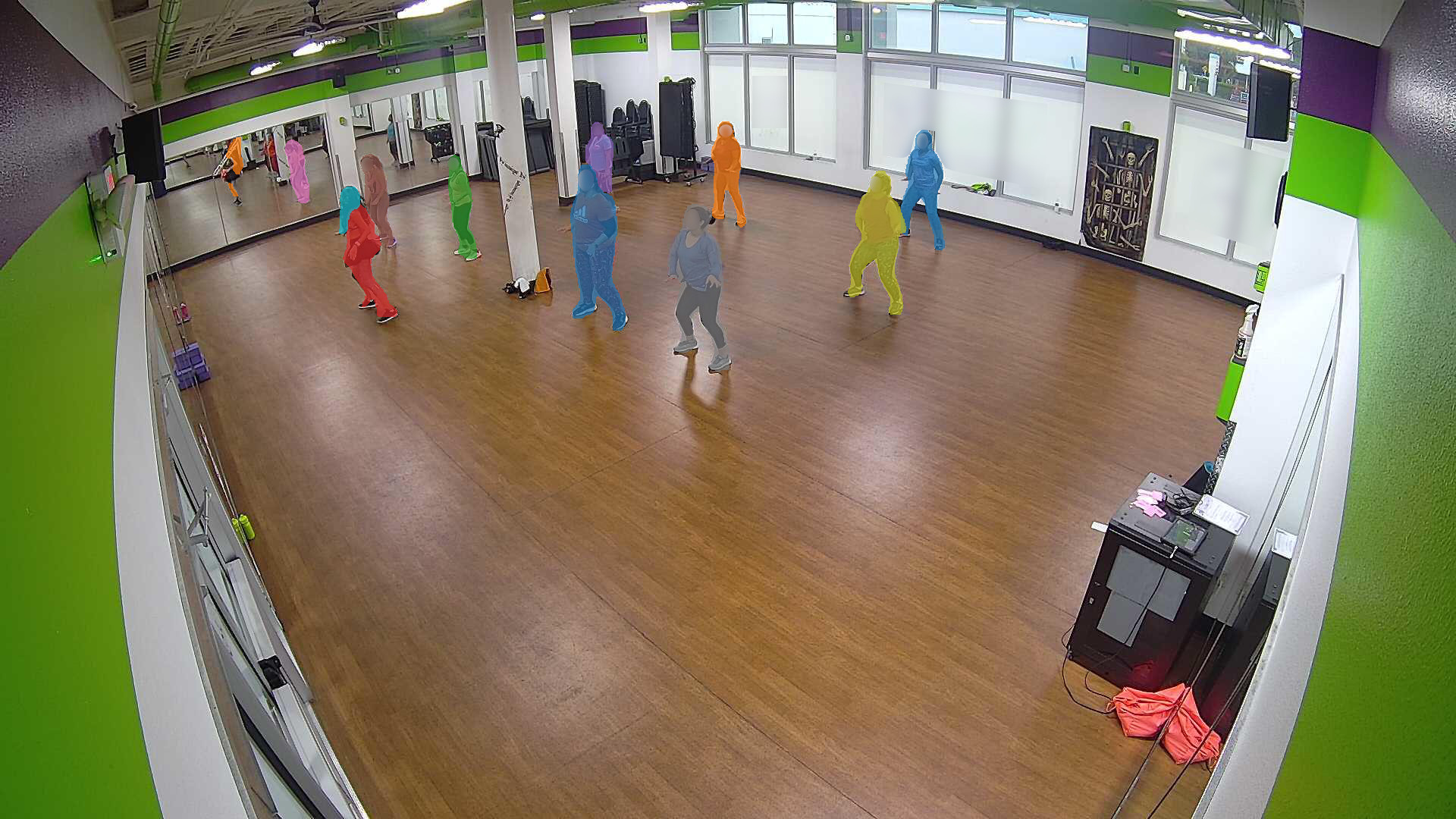}
	\includegraphics[width=0.19\textwidth]{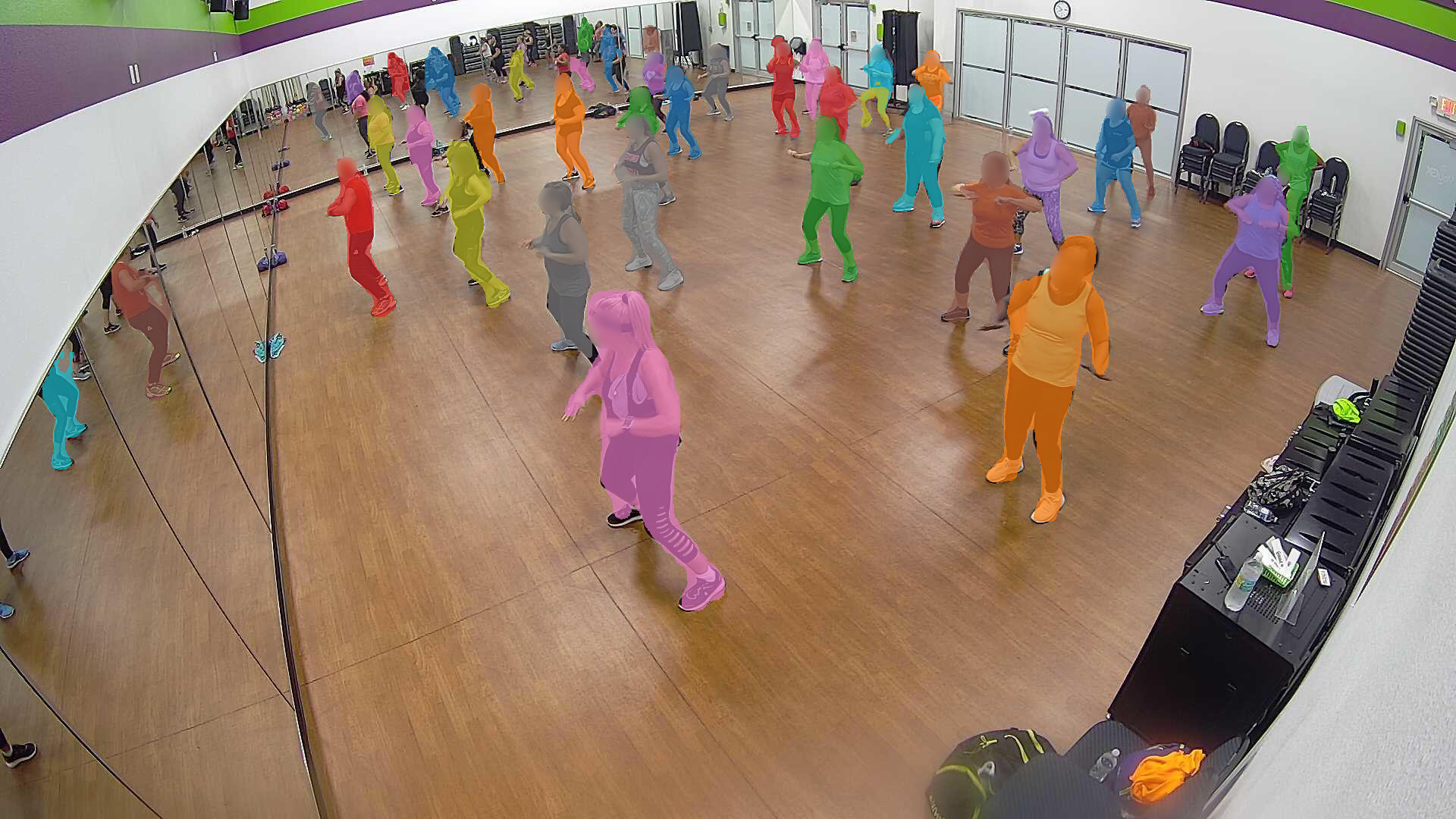}
	\includegraphics[width=0.19\textwidth]{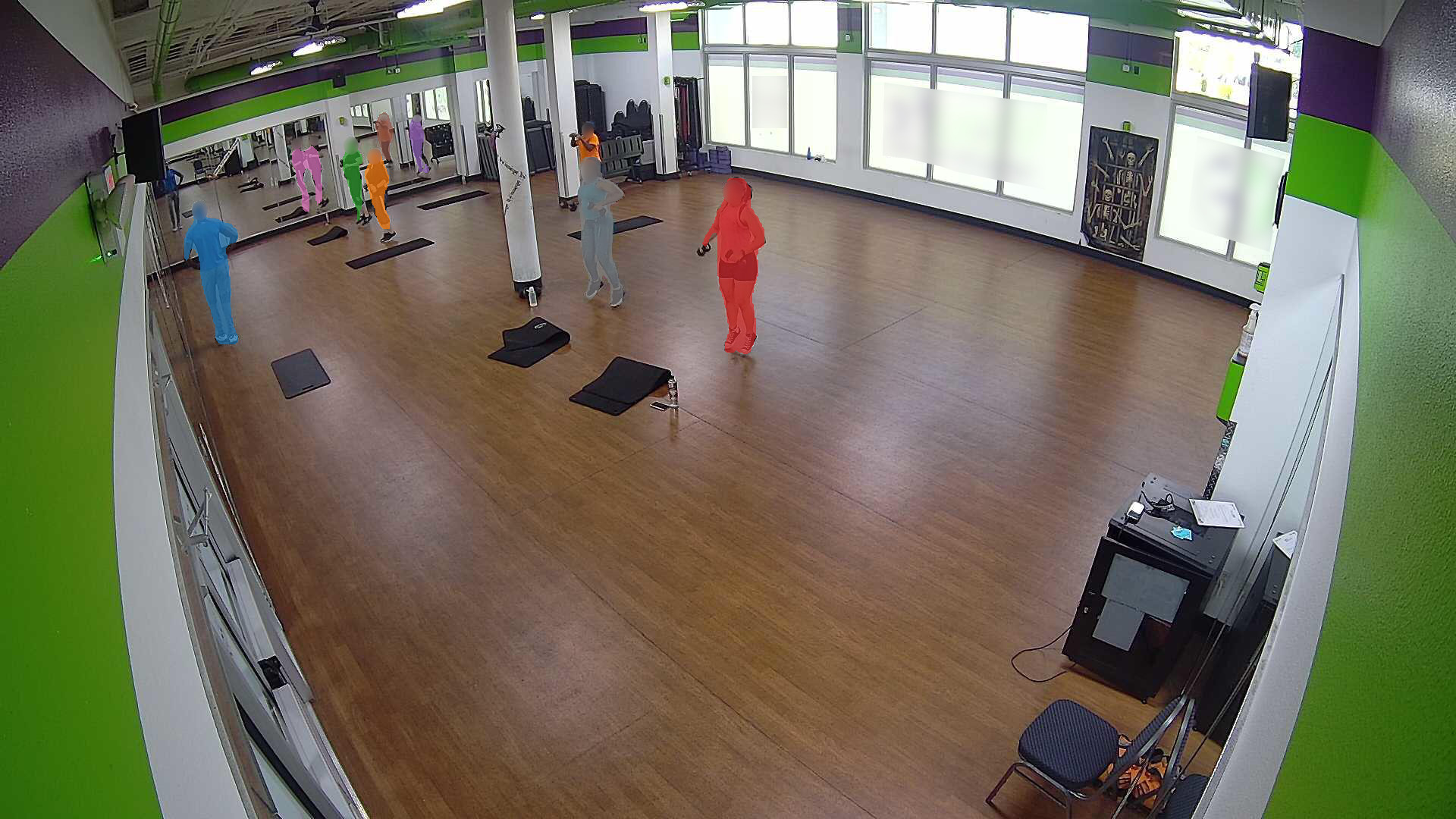}
	\includegraphics[width=0.19\textwidth]{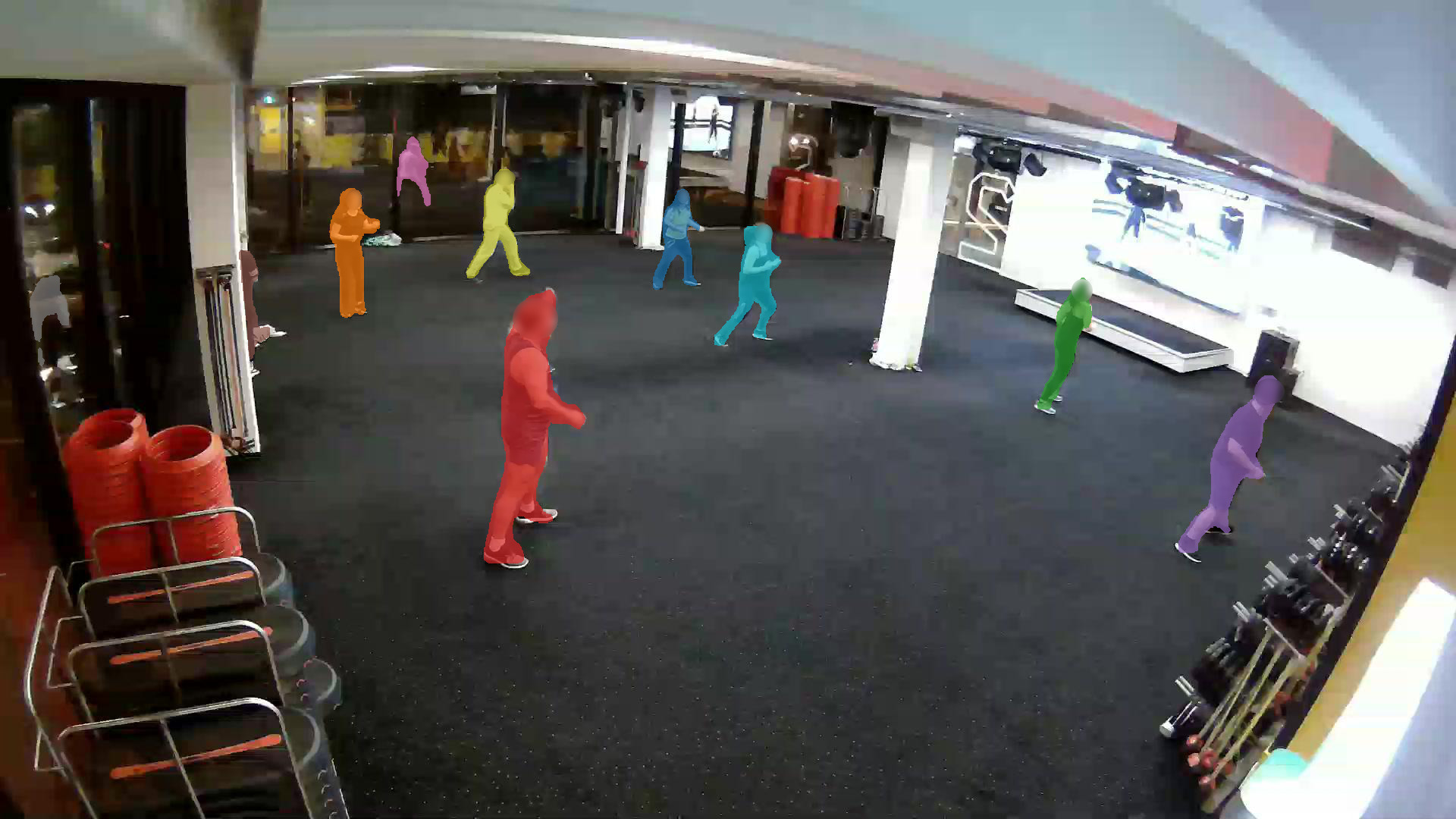}

	\includegraphics[width=0.19\textwidth]{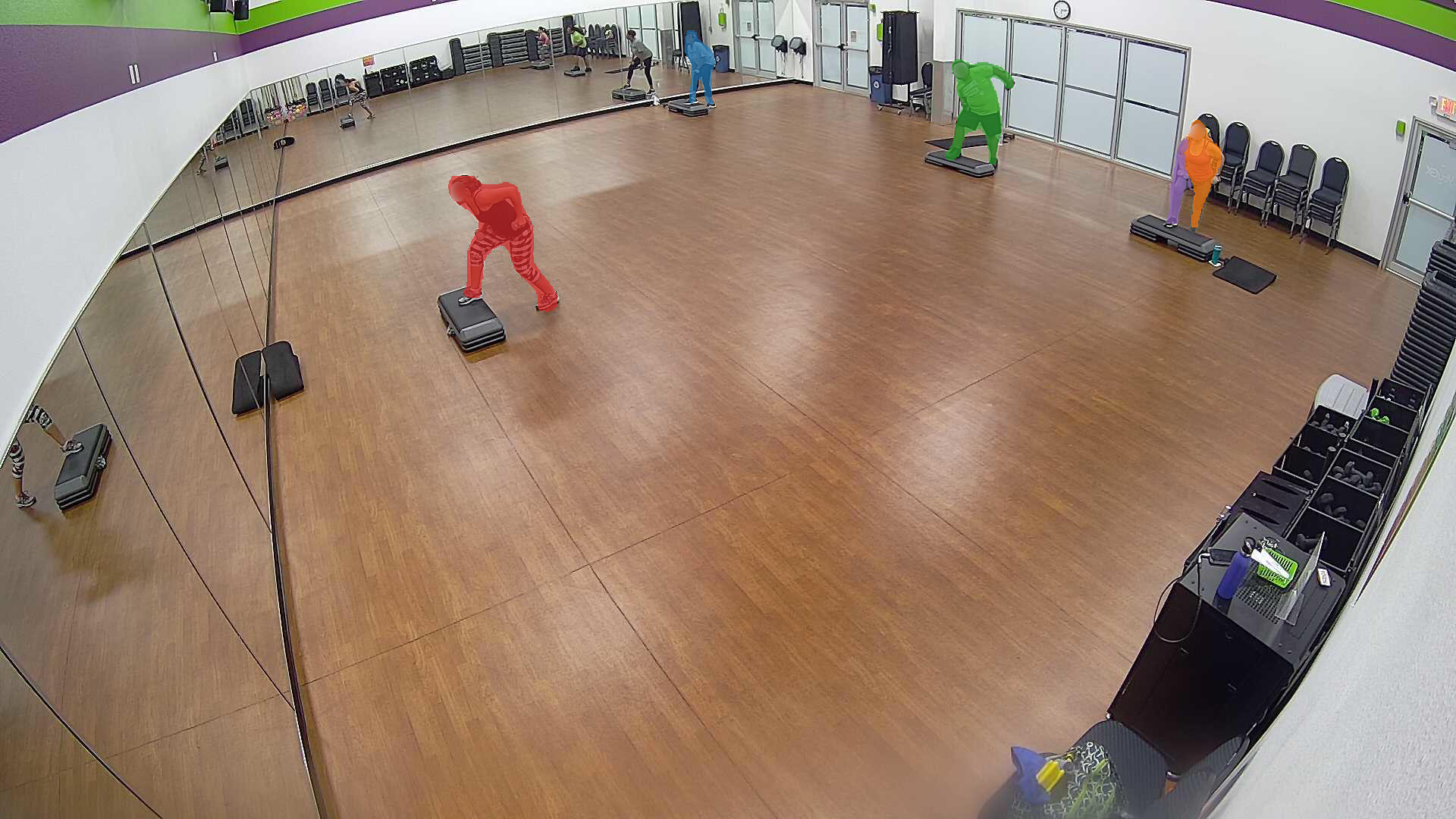}
	\includegraphics[width=0.19\textwidth]{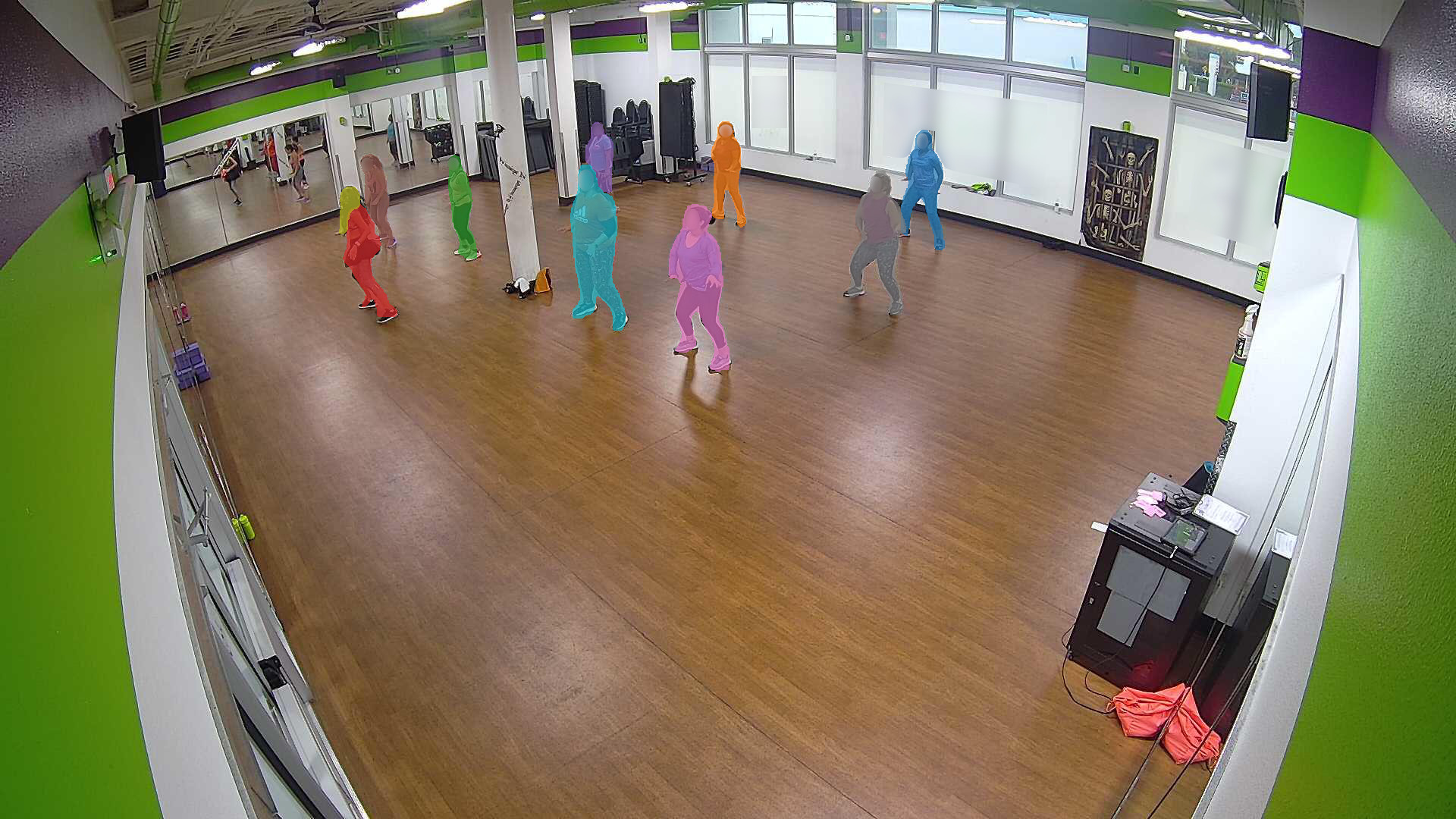}
	\includegraphics[width=0.19\textwidth]{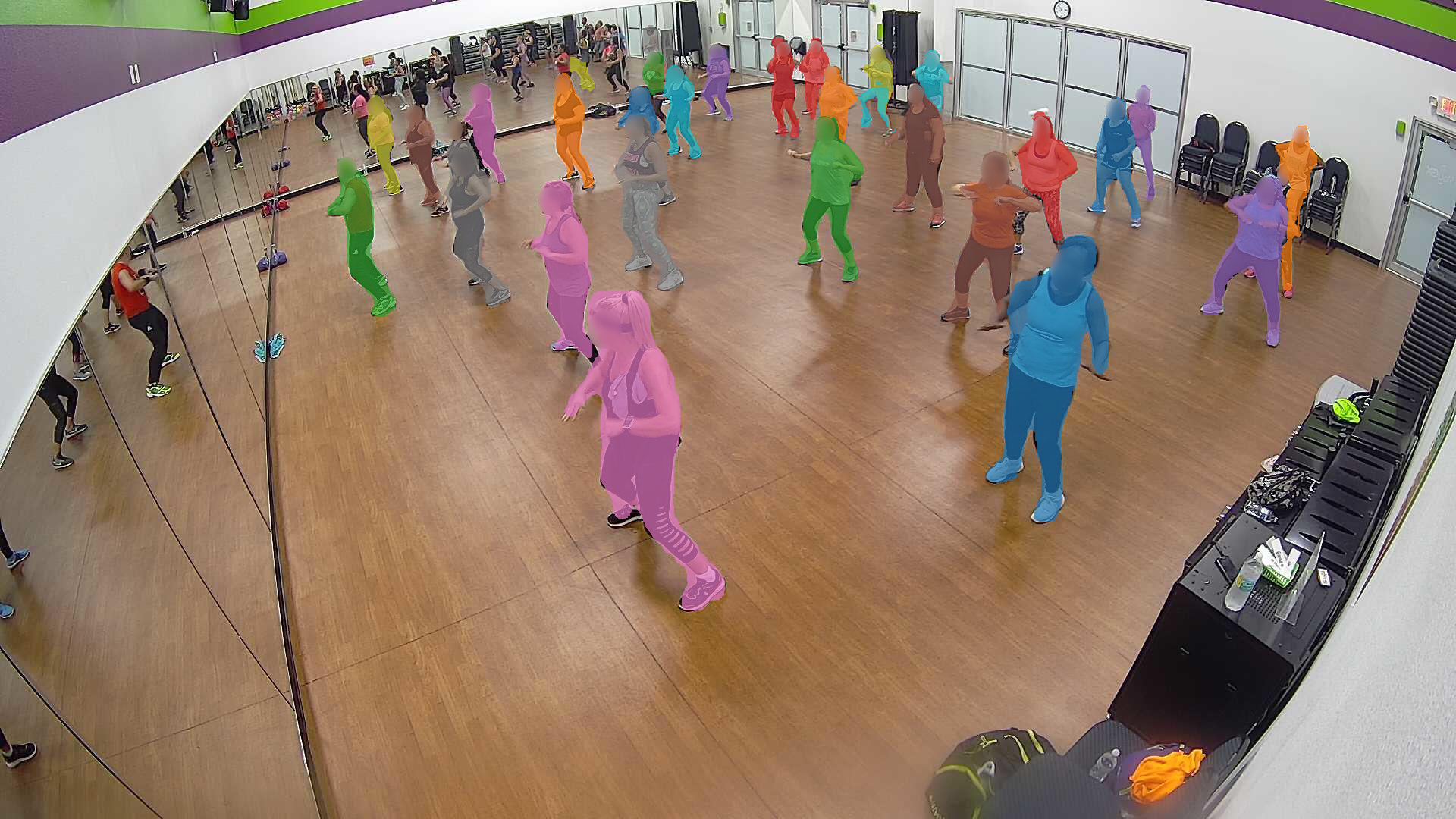}
	\includegraphics[width=0.19\textwidth]{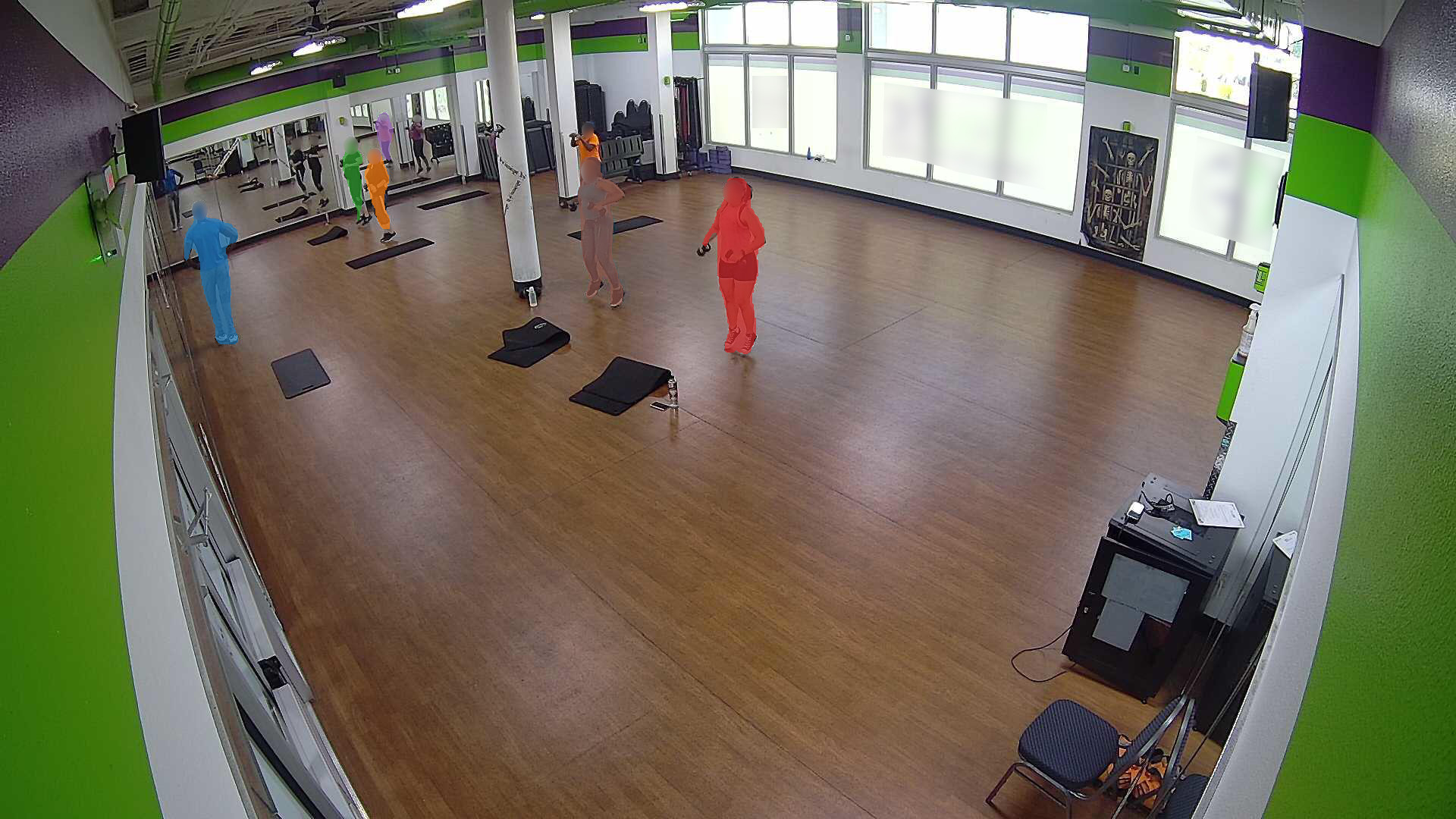}
	\includegraphics[width=0.19\textwidth]{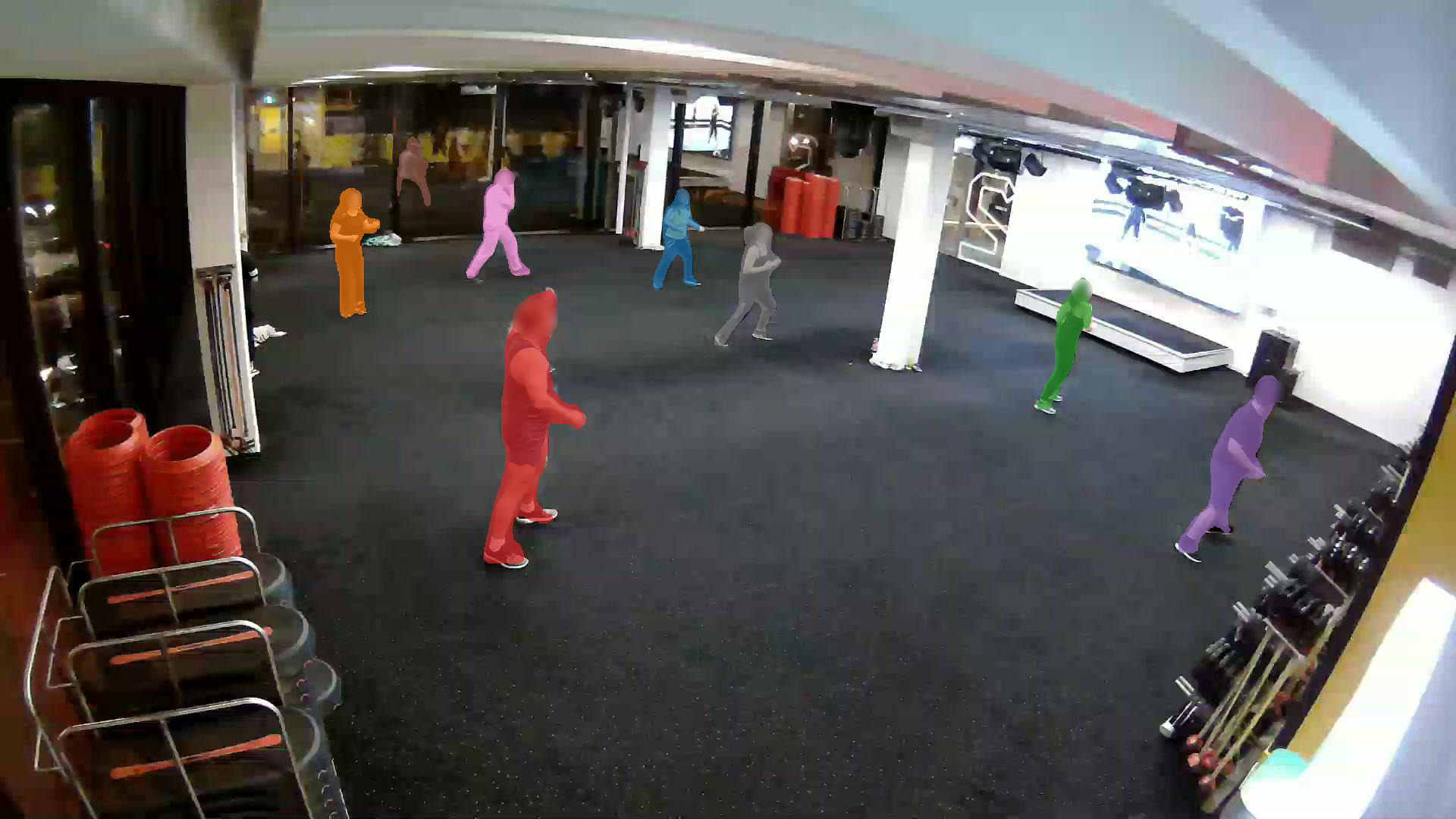}
	\caption{Pretrained on surveillance data without reflections. \textbf{Top Row}: Mask RCNN. \textbf{Bottom Row}: Joint.}
	\label{fig:orig-examples}
	\end{subfigure}
	\vspace{1em}
	\begin{subfigure}{\textwidth}
	\includegraphics[width=0.19\textwidth]{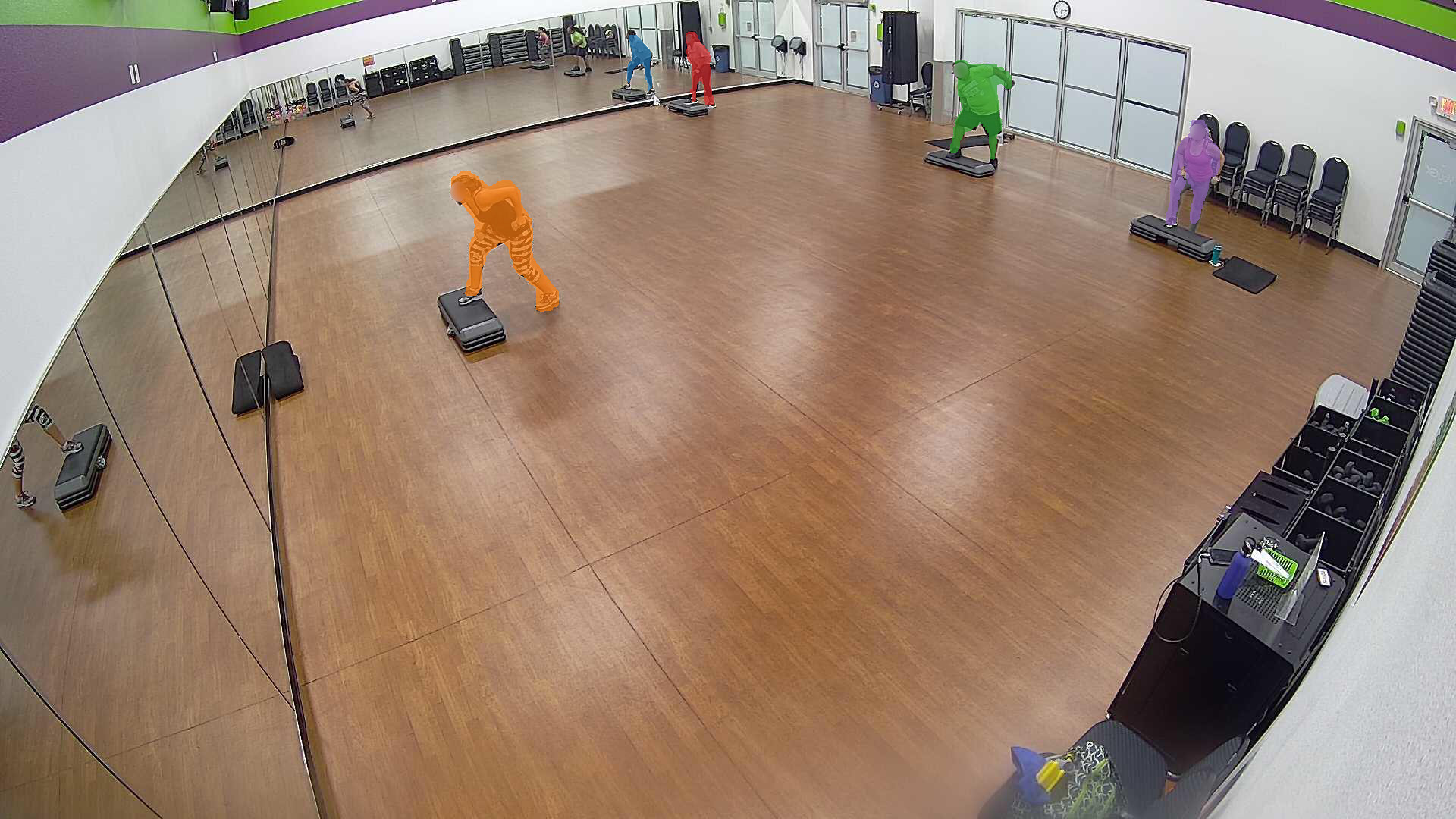}
	\includegraphics[width=0.19\textwidth]{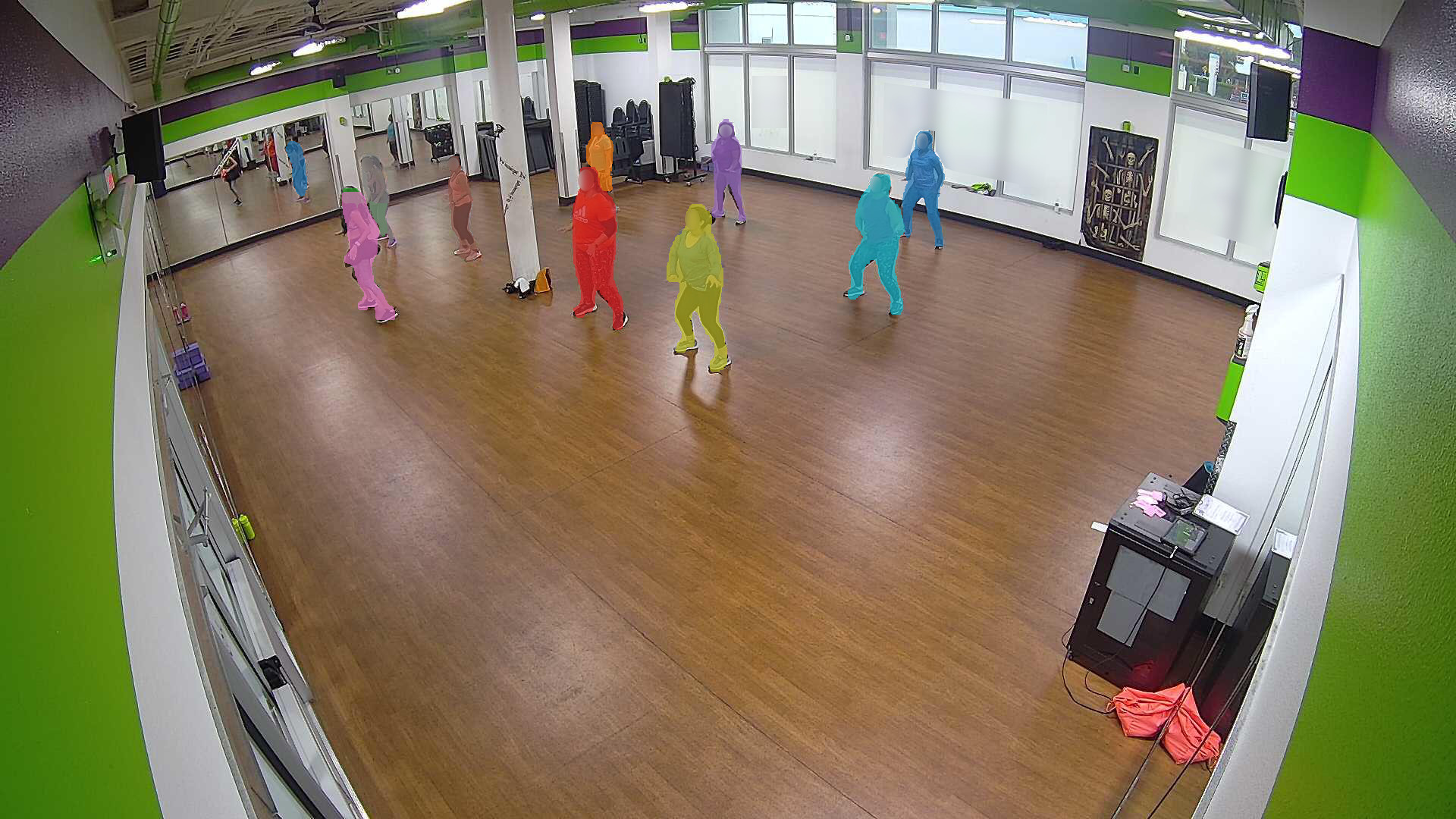}
	\includegraphics[width=0.19\textwidth]{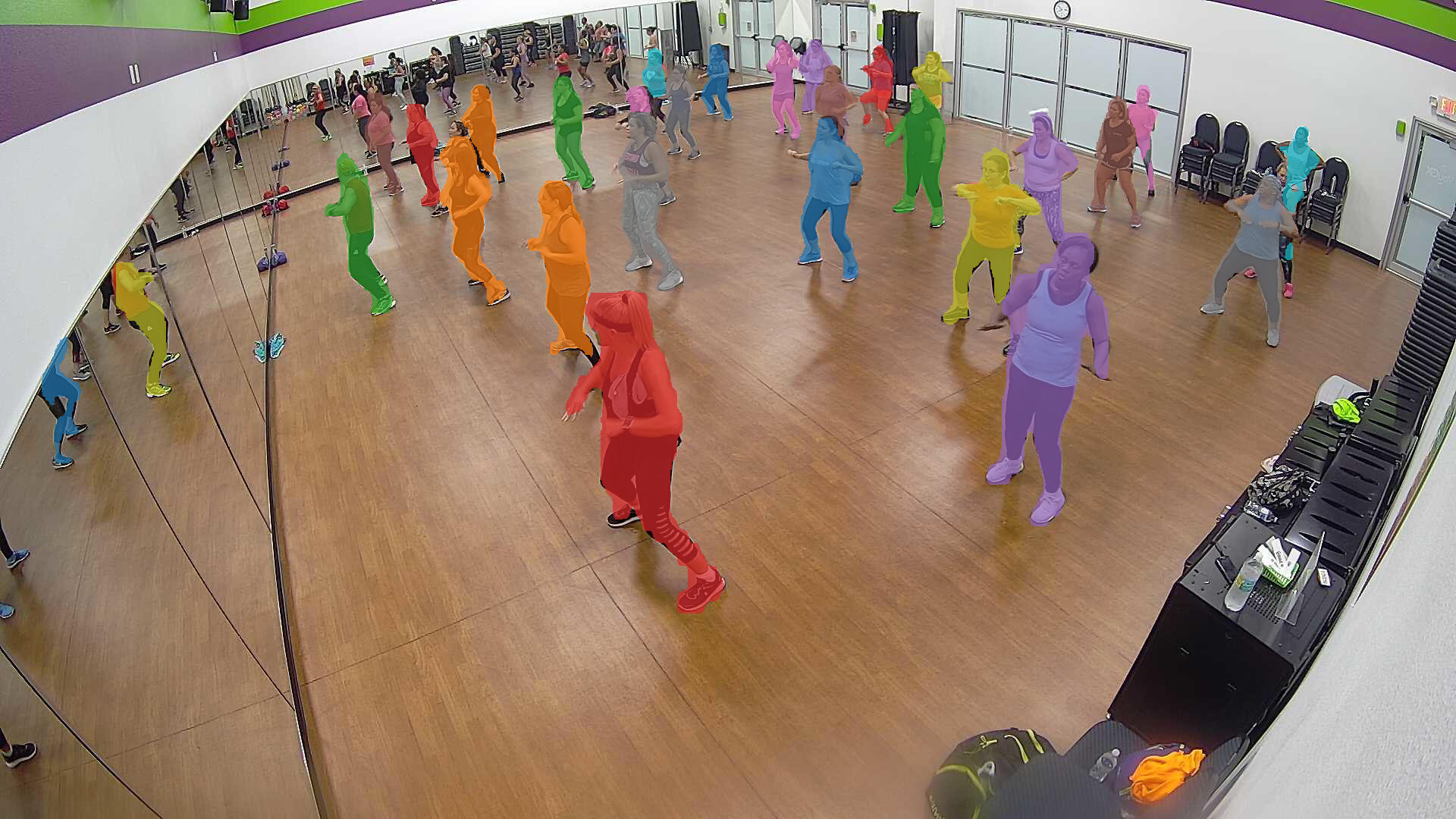}
	\includegraphics[width=0.19\textwidth]{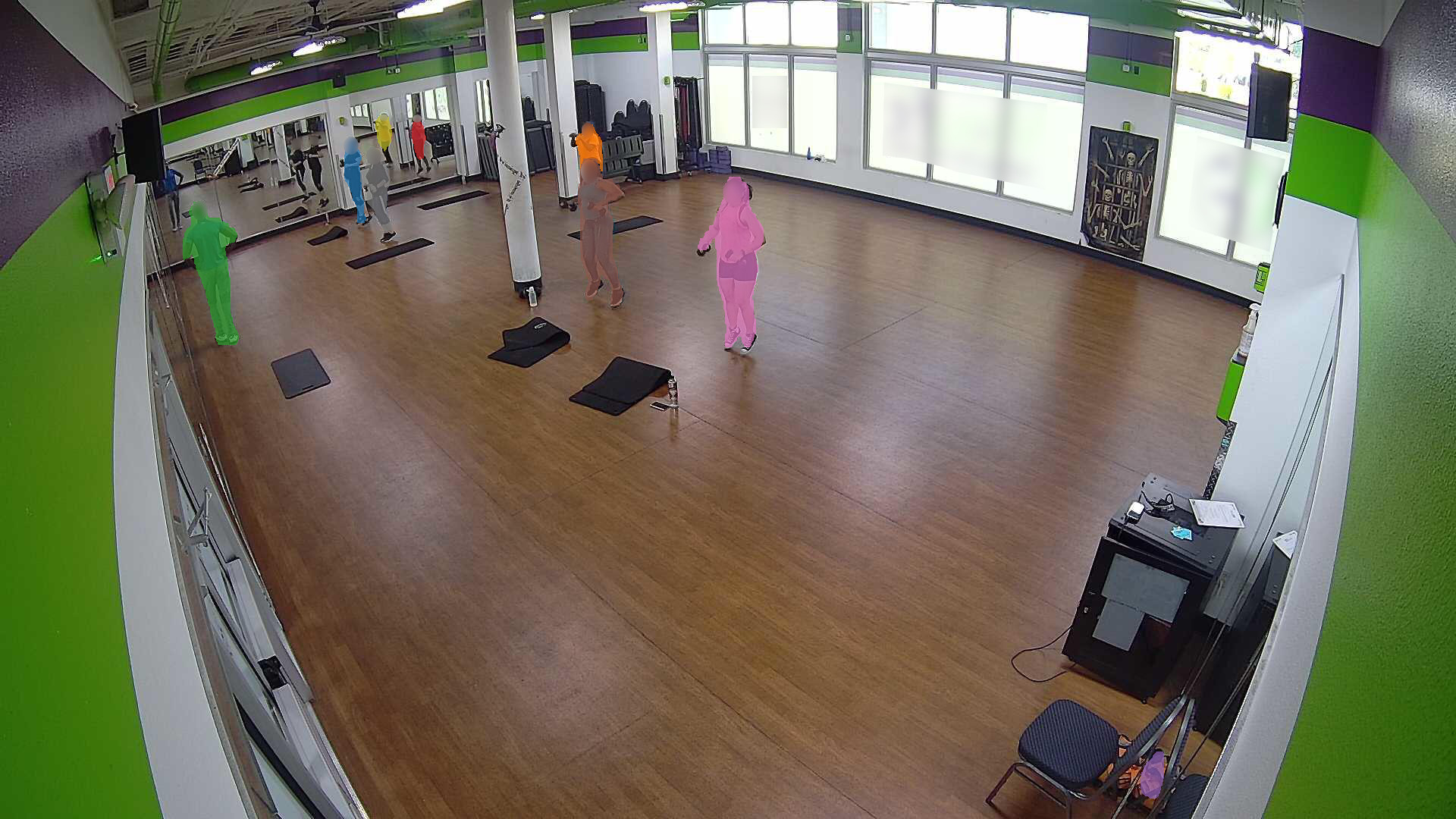}
	\includegraphics[width=0.19\textwidth]{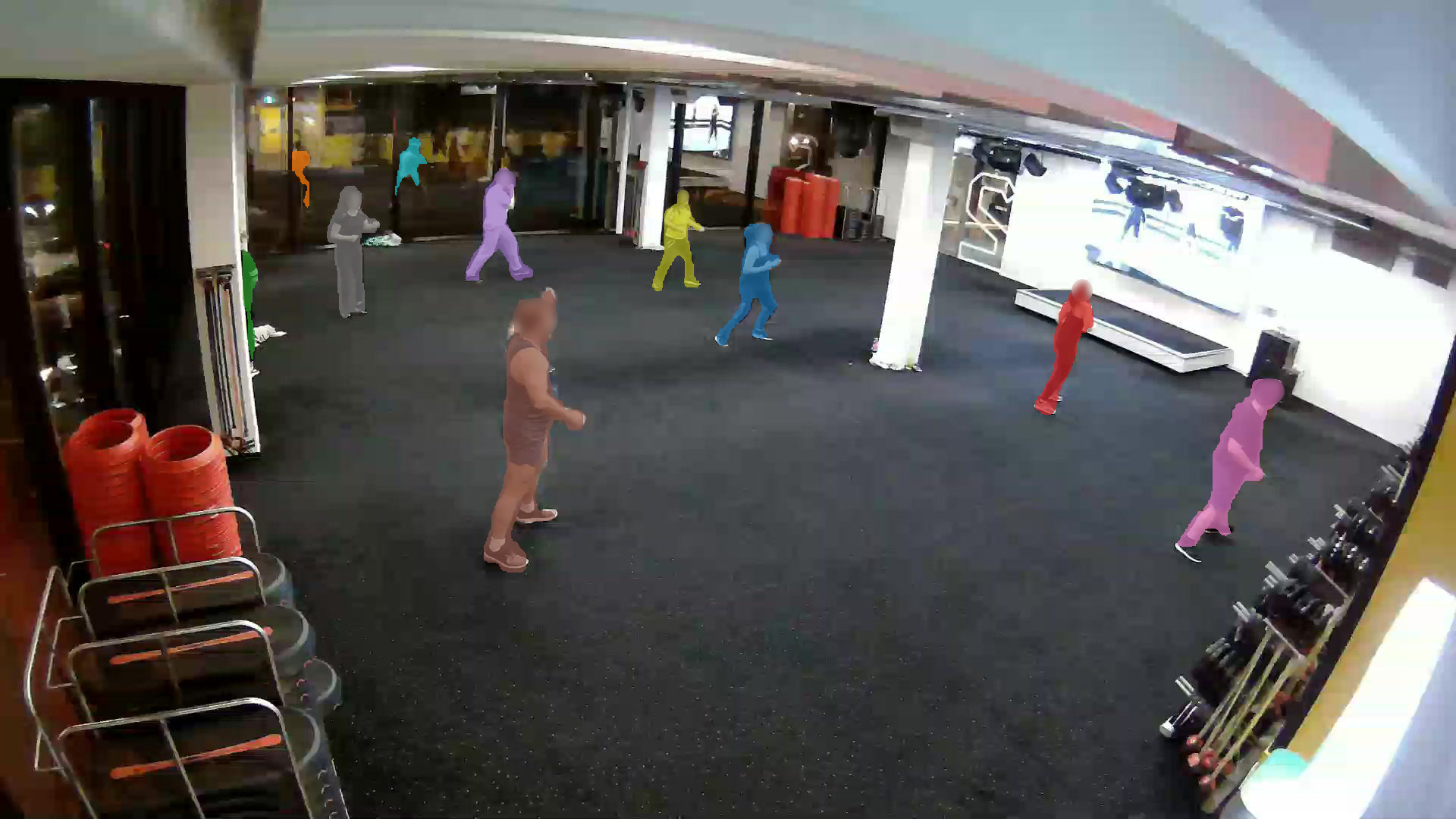}

	\includegraphics[width=0.19\textwidth]{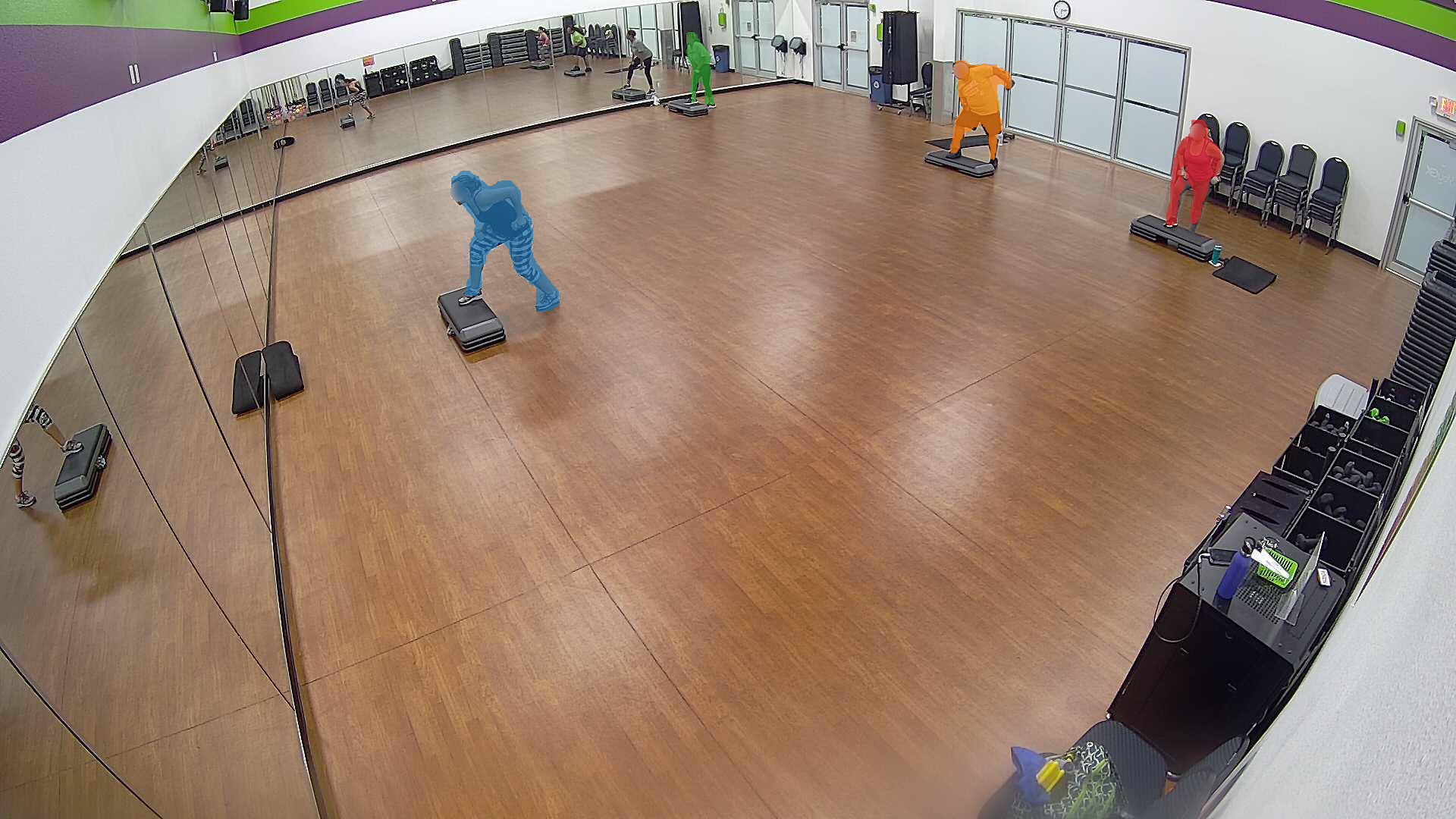}
	\includegraphics[width=0.19\textwidth]{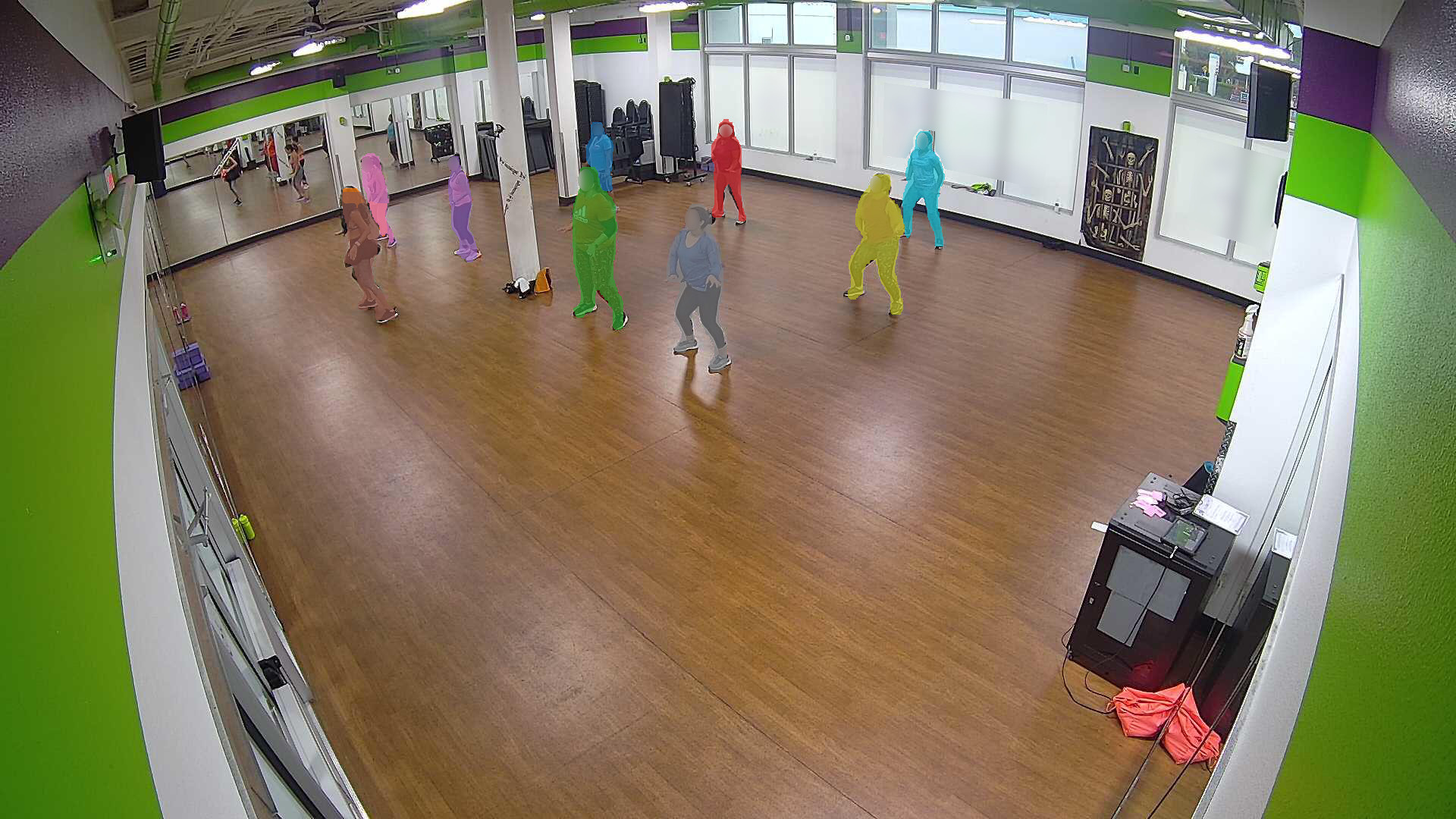}
	\includegraphics[width=0.19\textwidth]{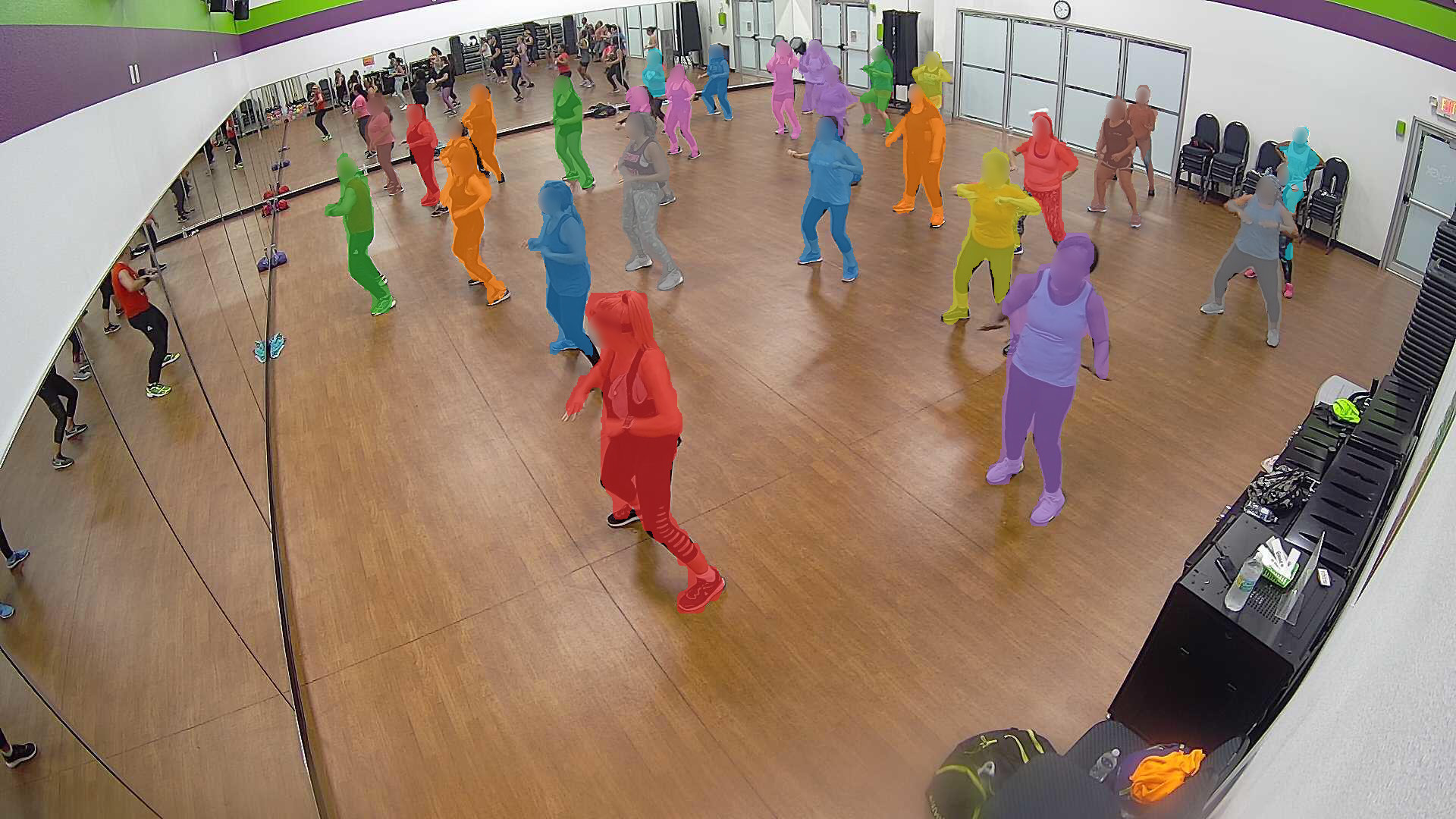}
	\includegraphics[width=0.19\textwidth]{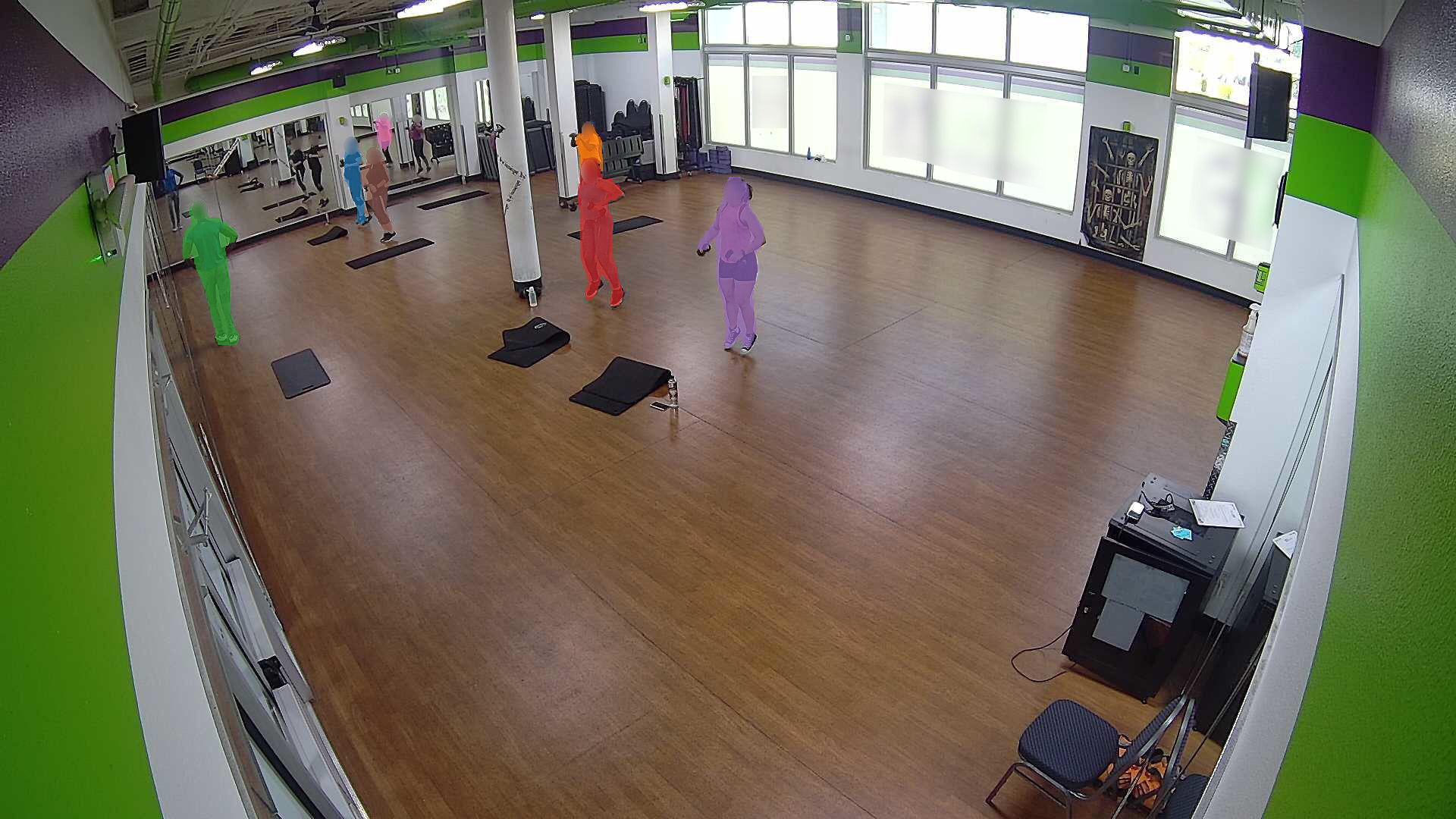}
	\includegraphics[width=0.19\textwidth]{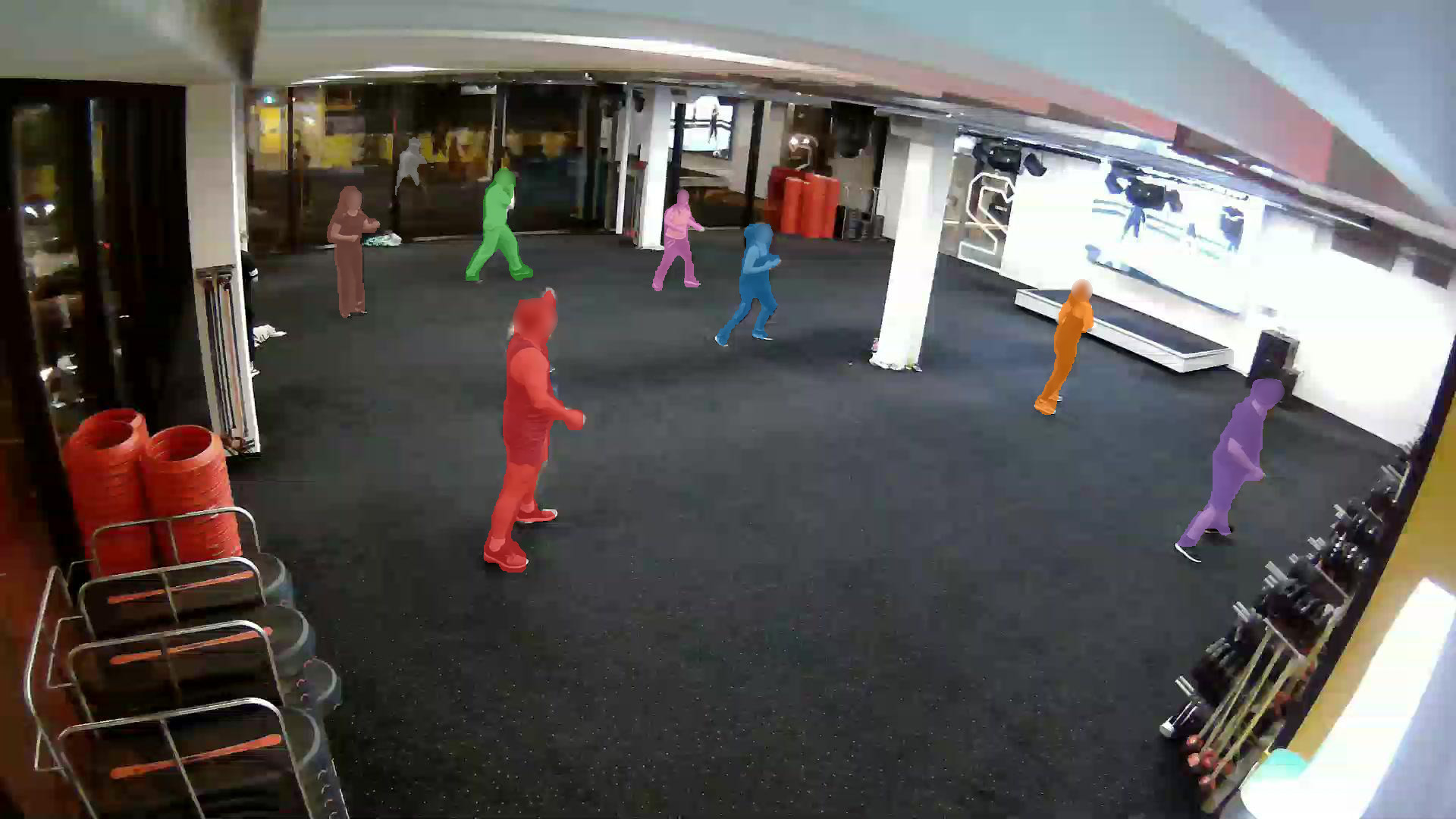}
	\caption{Pretrained and then finetuned on mirror reflections. \textbf{Top Row}: Mask RCNN. \textbf{Bottom Row}: Joint.}
	\label{fig:ft-examples}
	\vspace{1em}
	\end{subfigure}

	\caption{Example segmentations. Fusion can remove most reflections, even reflections from non-mirror surfaces such as windows.}
\end{figure}

Fusion of segmentations reduces false positives in both cases: pretrained and pretrained then finetuned. Figure~\ref{fig:orig-examples} shows several example images for the pretrained case, in which false positives are suppressed by fusion. As the fourth column shows, this does not yield a perfect result every time, but it generally helps and very seldom leads to the suppression of true positives. Failures to suppress reflections are generally due to the mirror frame being obstructed, as shown in the fourth column. Moreover, in the fifth column, we show that the joint approach is not limited to reflections in mirrors, but can also apply to reflections from other glossy surfaces, subject to having appropriately labelled data. In the original image a glossy window reflects the people present in the room, leading to false positives in the Mask RCNN result. Our procedure successfully filters one of these false positives.

\begin{figure}[h]
	\centering
	\begin{tabular}{c c c c c c c}
			 & FP   & FN  & Precision & Recall  & AP   & AR  \\
	\hline
		Mask RCNN    & 502  & 192 & 0.71 & 0.86 & 0.82 & 0.86\\
	     Joint    & 251  & 202 & 0.83 & 0.86 & 0.83 & 0.86 \\
	\hline
	\end{tabular}

	\includegraphics[width=0.40\textwidth]{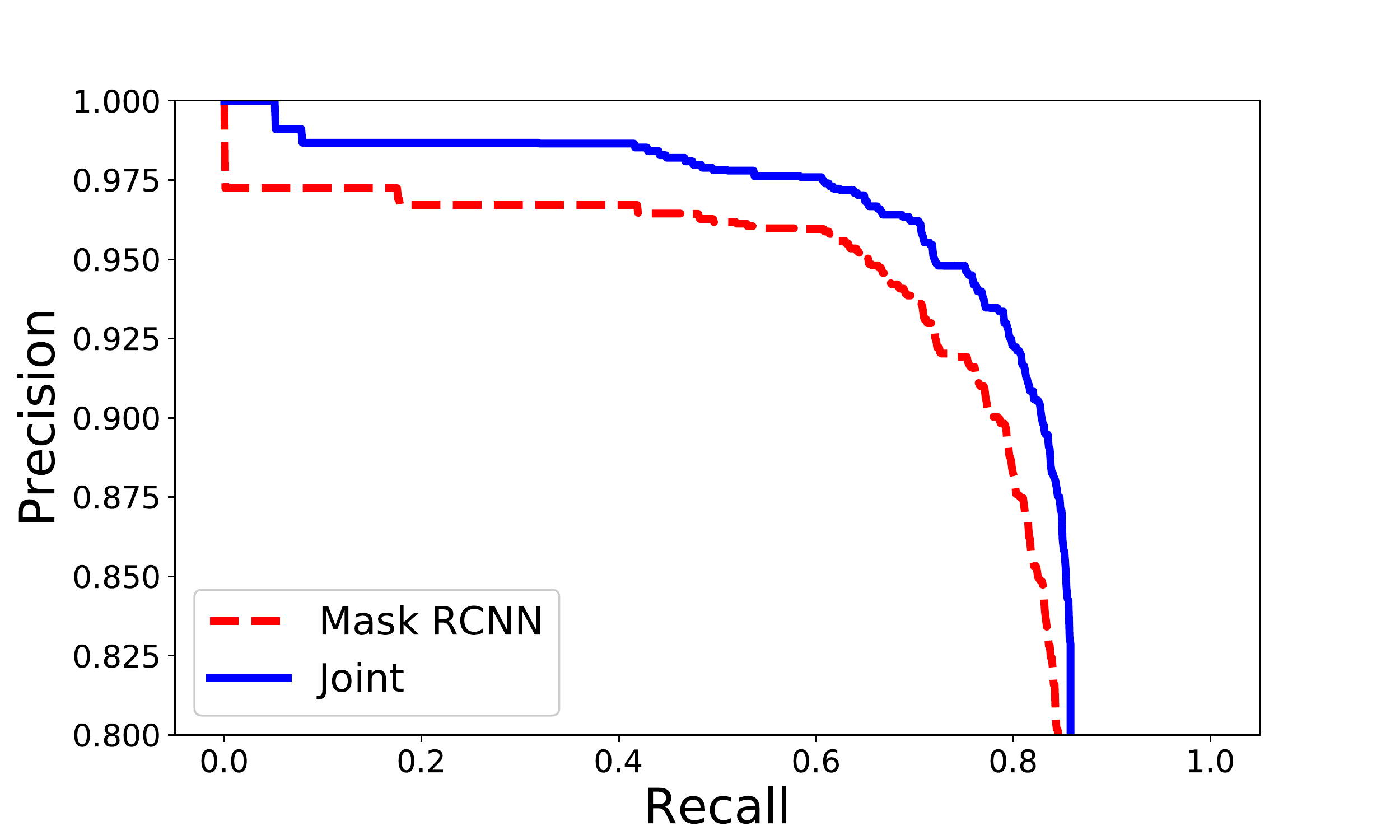}
	\includegraphics[width=0.40\textwidth]{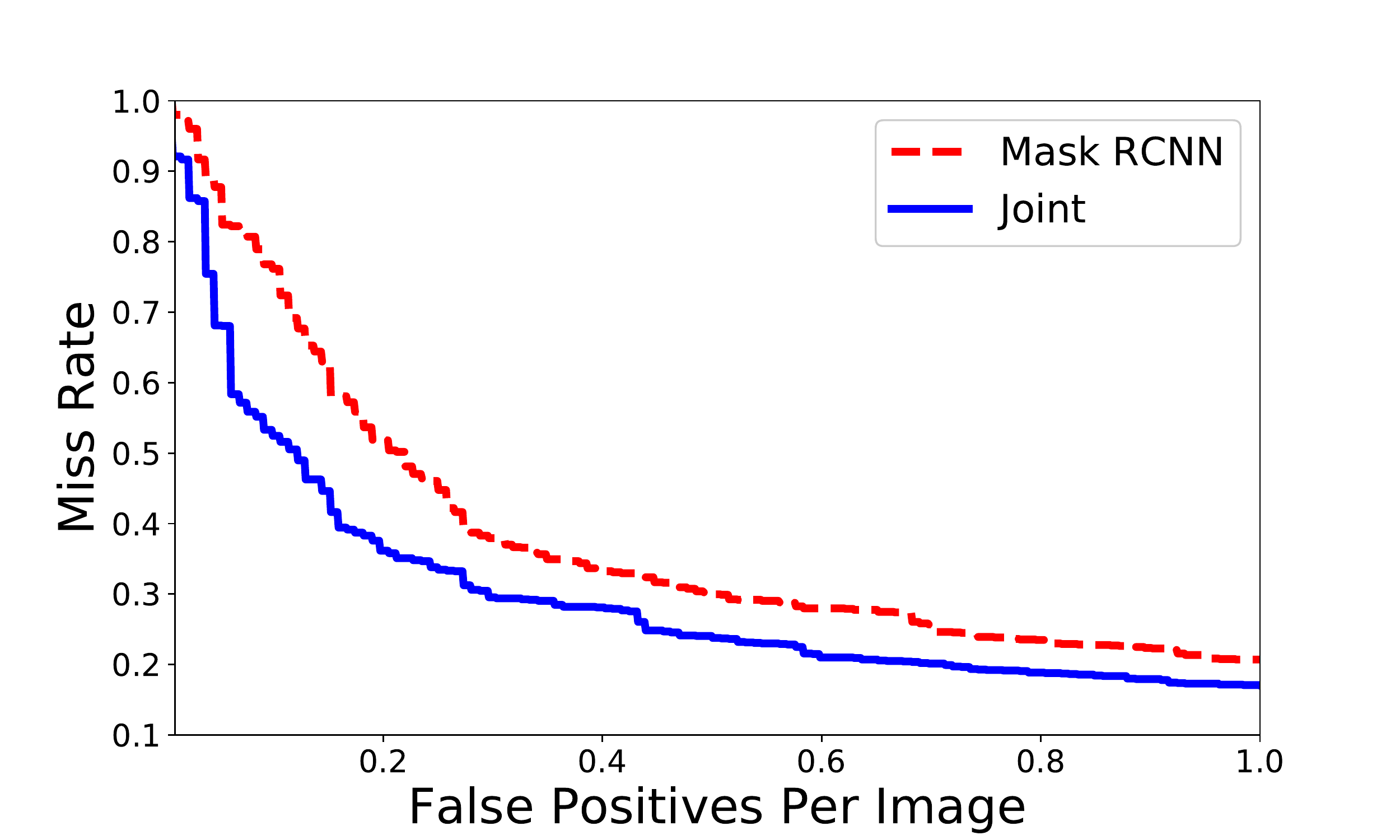}
	\caption{Performance metrics after pretraining on a surveillance dataset without mirror reflections, no finetuning. Here, the joint method provides an even larger performance boost, due to the severe effect of reflections. Despite this, the joint method restores performance close to that gotten from training on reflection-containing data.}
	\label{fig:orig-table}
\end{figure}

\begin{figure}[h]
	\centering
	\begin{tabular}{c c c c c c c}
			 & FP   & FN  & Precision & Recall  & AP   & AR  \\ 
	\hline
		Mask RCNN    & 300  & 188 & 0.80 & 0.87 & 0.80 & 0.87\\
	     Joint    & 227  & 196 & 0.84 & 0.86 & 0.80 & 0.86\\
	\hline
	\end{tabular}

	\includegraphics[width=0.40\textwidth]{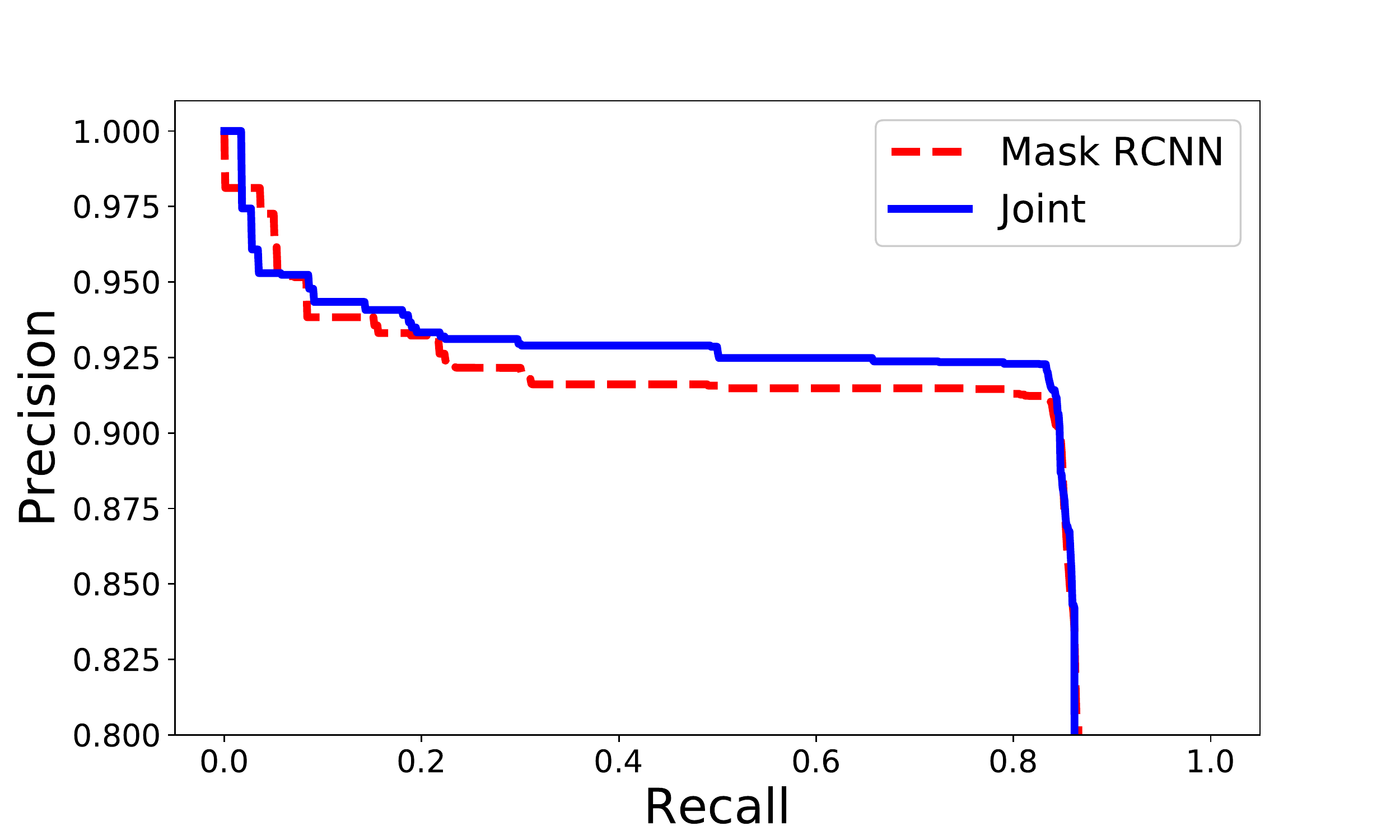}
	\includegraphics[width=0.40\textwidth]{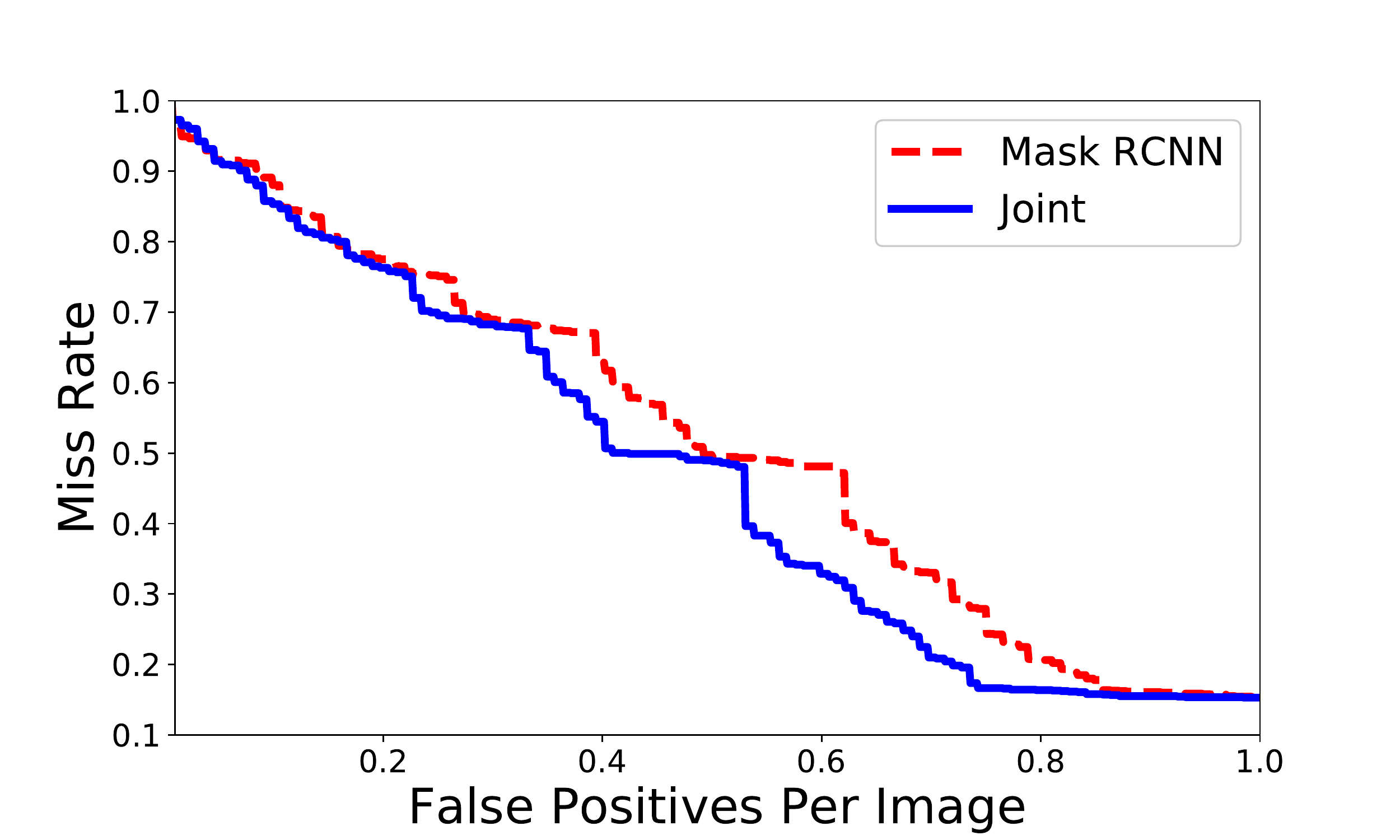}
	\caption{Performance metrics after pretraining and finetuning on a dataset containing mirror reflections. The joint method improves false positive count (and hence precision) without meaningfully affecting other performance metrics.}
	\label{fig:ft-table}
\end{figure}

These examples match the trend shown in the table of Figure~\ref{fig:orig-table}: Segmentation fusion halves the number of false positives from 502 to 251, hence increasing precision from 71\% to 83\%. Meanwhile the increase in false negatives is small, and the recall is practically unaffected (1\%). Notably, the average precision does not show much of this improvement, due to its being evaluated over a range of score thresholds. The curve of precision versus recall, however, reveals how joint segmentation improves performance over the most relevant regions of recall, leading to the observed in-dataset improvement in precision. This is matched well by the curve of miss rate versus false positives per image.

Figure~\ref{fig:ft-examples} shows example images for the case where the pretrained network is finetuned on images with mirror reflections. As expected, there are fewer false positives in the Mask RCNN results, as finetuning has trained the network to discard some of them. Still, some false positives remain, and fusion continues to provide a benefit. This is not entirely surprising, but emphasises how the semantic segmentation can inject information that the instance segmentation struggles to use, and shows that fusion is useful where finetuning fails to remove the mirror reflection false positives.
As in the pretrained case, summary measures such as the average precision de-emphasise the significant improvement in in-dataset precision, but improvement is clear in plots of precision versus recall and miss rate versus false positives per image. These examples match the trend shown in the table of Figure~\ref{fig:ft-table}: In the pretrained+finetuned network fusion greatly reduces the false positives in the dataset, while producing only marginal changes in the number of false negatives. Joint segmentation thus boosts the precision from 80\% to 84\%, with other statistics practically unaffected.

Perhaps surprisingly, finetuning Mask RCNN \emph{without} fusion performed slightly worse than simply using non-finetuned Mask RCNN \emph{with} fusion (precision 80\% versus 83\%, recall 87\% versus 86\%). The best option, of course, is to use fusion \emph{and} finetune Mask RCNN -- but this disparity shows fusion not only offers additional benefits beyond what can be provided by finetuning of the instance segmentation, but also how incorporating semantic segmentation can even reduce or remove the need for finetuning of the instance branch.

\section{Conclusion}
We have shown how joint semantic and instance segmentation, even using a straightforward heuristic approach, can dramatically reduce the incidence of false positives due to mirror reflections by reintroducing broader context to instance segmentations. We have also demonstrated in detail how standard two-stage instance segmentation, such as Mask RCNN, is particularly vulnerable to these false positives; and we have shown how a joint approach inspired by panoptic segmentation reintroduces image context to address this vulnerability.

A clear direction for future work would be to approach the joint segmentation task in a more rigorous, unified way, as some authors have already begun to do~\cite{kirillov2019panoptic,li2018attention,de2018panoptic} -- and to explore how these unified approaches cope with detecting false positives due to mirror reflections. Moreover, we might hope to see mutual improvement between semantic and instance segmentation in other settings: for example, this has already been noted as a direction for research in small object detection~\cite{bell2016inside}, and panoptic segmentation may be particularly relevant there. More generally, initial results in end-to-end panoptic segmentation~\cite{kirillov2019panoptic} and integrating a semantic branch into instance segmentation~\cite{chen2019hybrid} have shown mutual improvement between the two tasks in the COCO and Cityscapes datasets.

Nevertheless, our work shows how joint semantic and instance segmentation, often called \textit{panoptic} segmentation, is already valuable in practical computer vision applications. Our results support the growing excitement around panoptic segmentation, and point to the potential of unifying scene parsing and instance segmentation.

\bibliography{bib}

\end{document}